\documentclass[pdflatex,sn-mathphys-num]{sn-jnl}

\usepackage{graphicx}%
\usepackage{multirow}%
\usepackage[font=scriptsize]{caption} 
\usepackage{amsmath,amssymb,amsfonts}%
\usepackage{amsthm}%
\usepackage{mathrsfs}%
\usepackage[title]{appendix}%
\usepackage{textcomp}%
\usepackage{manyfoot}%
\usepackage{booktabs}%
\usepackage{algorithm}%
\usepackage{algorithmicx}%
\usepackage{algpseudocode}%
\usepackage{listings}
\usepackage{todonotes}
\usepackage{parskip}
\usepackage{booktabs}
\usepackage{xcolor}
\definecolor{codegreen}{rgb}{0,0.6,0}
\definecolor{codegray}{rgb}{0.5,0.5,0.5}
\definecolor{codepurple}{rgb}{0.58,0,0.82}
\definecolor{backcolour}{rgb}{0.95,0.95,0.92}
\definecolor{metablue}{HTML}{0064E0}

\lstdefinestyle{mystyle}{
    commentstyle=\color{codegreen},
    keywordstyle=\color{magenta},
    numberstyle=\tiny\color{codegray},
    stringstyle=\color{codepurple},
    basicstyle=\ttfamily\footnotesize,
    breakatwhitespace=false,         
    breaklines=true,                 
    captionpos=b,                    
    keepspaces=true,                 
    numbers=left,                    
    numbersep=5pt,                  
    showspaces=false,                
    showstringspaces=false,
    showtabs=false,                  
    tabsize=2
}

\begin{document}

\title[]{\textbf{Graph Deep Learning for Intracranial Aneurysm Blood Flow Simulation and Risk Assessment}}

\author[1]{\fnm{Paul} \sur{Garnier}}

\author[1]{\fnm{Pablo} \sur{Jeken-Rico}}

\author[1]{\fnm{Vincent} \sur{Lannelongue}}

\author[1]{\fnm{Chiara} \sur{Faitini}}
\author[1]{\fnm{Aurèle} \sur{Goetz}}
\author[1]{\fnm{Lea} \sur{Chanvillard}}
\author[1]{\fnm{Ramy} \sur{Nemer}}
\author[1]{\fnm{Jonathan} \sur{Viquerat}}

\author[1]{\fnm{Ugo} \sur{Pelissier}}
\author[1]{\fnm{Philippe} \sur{Meliga}}

\author[2]{\fnm{Jacques} \sur{Sédat}}
\author[3]{\fnm{Thomas} \sur{Liebig}}
\author[2]{\fnm{Yves} \sur{Chau}}
\author*[1]{\fnm{Elie} \sur{Hachem}}\email{elie.hachem@minesparis.psl.eu}

\affil[1]{\orgdiv{Centre for Material Forming (CEMEF)}, \orgname{Mines Paris - PSL University, CNRS UMR 7635}, \orgaddress{\city{Sophia Antipolis}, \postcode{06904}, \country{France}}}

\affil[2]{\orgdiv{Interventional Neuroradiology Department}, \orgname{Nice University Hospital}, \orgaddress{\city{Nice}, \postcode{06100}, \country{France}}}

\affil[3]{\orgdiv{Department of Neuroradiology}, \orgname{University Hospital Munich (LMU)}, \orgaddress{\city{Munich}, \country{Germany}}}


\abstract{
Intracranial aneurysms remain a major cause of neurological morbidity and mortality worldwide, where rupture risk is tightly coupled to local hemodynamics particularly wall shear stress and oscillatory shear index. Conventional computational fluid dynamics simulations provide accurate insights but are prohibitively slow and require specialized expertise. Clinical imaging alternatives such as 4D Flow MRI offer direct in‑vivo measurements, yet their spatial resolution remains insufficient to capture the fine‑scale shear patterns that drive endothelial remodeling and rupture risk while being extremely impractical and expensive. 

We present a graph neural network surrogate model that bridges this gap by reproducing full‑field hemodynamics directly from vascular geometries in less than one minute per cardiac cycle. Trained on a comprehensive dataset of high‑fidelity simulations of patient‑specific aneurysms, our architecture combines graph transformers with autoregressive predictions to accurately simulate blood flow, wall shear stress, and oscillatory shear index. The model generalizes across unseen patient geometries and inflow conditions without mesh‑specific calibration. Beyond accelerating simulation, our framework establishes the foundation for clinically interpretable hemodynamic prediction. By enabling near real‑time inference integrated with existing imaging pipelines, it allows direct comparison with hospital phase‑diagram assessments and extends them with physically grounded, high‑resolution flow fields.

This work transforms high-fidelity simulations from an expert-only research tool into a deployable, data-driven decision support system. Our full pipeline delivers high-resolution hemodynamic predictions within minutes of patient imaging, without requiring computational specialists, marking a step-change toward real-time, bedside aneurysm analysis.
}


\maketitle

\section{Introduction}\label{sec:introduction}

Intracranial aneurysms (IA) affect an estimated 2--3\% of the adult population\,\cite{Rinkel1998, Etminan2016, Etminan2019}, and often remain clinically silent until rupture, which results in death or severe neurological disability in over 75\% of cases\,\cite{Molyneux2005, Etminan2019}. These vascular malformations, commonly found around the Circle of Willis, evolve through complex cycles of growth, remodeling, and potential rupture\,\cite{Etminan2016}. Although the precise biological mechanisms remain partially understood, cerebral hemodynamics, genetic predisposition, and local wall pathology all contribute to their development and progression\,\cite{Vlak2011, Hindenes2023, Etminan2016}.

To prevent rupture, current treatment strategies aim to either exclude the aneurysm from circulation or limit the blood flow circulation inside the bulge. Surgical clipping, endovascular coiling or flow diversion are routinely employed\,\cite{Molyneux2005}, yet each intervention carries risk, and treatment decisions remain controversial, particularly for unruptured aneurysms\,\cite{Kulcsar2011, Thompson2015, Peschillo2016}. A major challenge is the lack of robust, non-invasive risk assessment tools to distinguish high-risk from stable aneurysms\,\cite{Pagiola2020, Berg2019}. A growing body of evidence suggests that biomechanical indicators, especially Wall Shear Stress (WSS) and Oscillatory Shear Index (OSI), play a central role in aneurysm initiation, progression and more importantly: rupture \cite{Meng2014, Berg2016, Sarrami-Foroushani2021}. Elevated WSS can damage endothelial cells and initiate aneurysm formation via delamination of the internal elastic lamina\,\cite{Meng2014, Etminan2016}, while regions of low and oscillatory WSS promote inflammatory remodelling pathway, commonly leading to plaque buildup \cite{Meng2014}.

Computational Fluid Dynamics (CFD) has become a critical tool for modeling patient-specific hemodynamics, allowing high-resolution evaluation of WSS, OSI, and other flow-derived biomarkers. However, its clinical adoption remains limited due to the high computational cost, the need for expert intervention at every pipeline step, and inconsistent reproducibility across modeling assumptions\,\cite{Valen-Sendstad2018, voss2019, Berg2019rev}. More recently, 4D flow MRI has been proposed as an alternative, offering in-vivo blood velocity fields without numerical simulation\,\cite{Urschel2021}. Yet, its spatial resolution is often insufficient for resolving fine-scale structures critical to aneurysm rupture, and it remains confined to specialized centers with advanced MRI hardware.

In parallel, machine learning (ML) is revolutionizing computational physics. Initial ML models for fluid simulation relied on convolutional architectures applied to structured grids\,\cite{Tompson2016, Thuerey2018, Chen2019}, with limited scalability to 3D anatomical structures. Graph Neural Networks (GNNs), which operate natively on unstructured meshes, have since emerged as a powerful alternative\,\cite{pfaff2021learning, sanchezgonzalez2020learning}. GNN-based surrogate models can learn to predict the evolution of fluid fields directly from geometry and boundary conditions, enabling real-time simulation once trained\,\cite{Suk_2024, Libao2023}. Most GNN methods applied to fluid dynamics now use variants of multigrid architectures~\cite{John2002}, where message passing is applied at different scales~\cite{lin2024upsamplingonlyadaptivemeshbasedgnn, Fortunato2022, lino2021simulating, Yang2022amgnet, taghibakhshi2023mggnn, cao2023efficientlearningmeshbasedphysical}.

Our work builds on this paradigm shift. We first built a dataset comprises over 100 synthetically standardized geometries mimicking real aneurysm shapes extracted from the INTRA database\,\cite{Goetz2024b}, forming the pre-training corpus. We fine-tune the model on 13 real patient-specific aneurysms simulated under six distinct physiological inflow waveforms, leading to more than 40 cases, substantially increasing the diversity of anatomical and boundary condition scenarios. For final validation, we adopt the five aneurysms from the MATCH challenge\,\cite{voss2019}, which feature out-of-distribution patterns in both geometry and hemodynamics. 

The full model architecture is a graph-based transformer\,\cite{Vaswani2017} enhanced with an augmented adjacency matrix\,\cite{garnier2024transformer}, enabling long-range attention across unstructured meshes of up to 4 million elements. Each node carries 15 variables, including velocity, acceleration, spatial coordinates, node type, and boundary flow statistics, representing a single snapshot in a typical cardiac cycle. The model operates in an autoregressive fashion, producing $\Delta t = 0.01s$ one-step predictions to reconstruct the entire cardiac cycle.

\begin{figure}[!ht]
	\centering
	\includegraphics[width=1\textwidth]{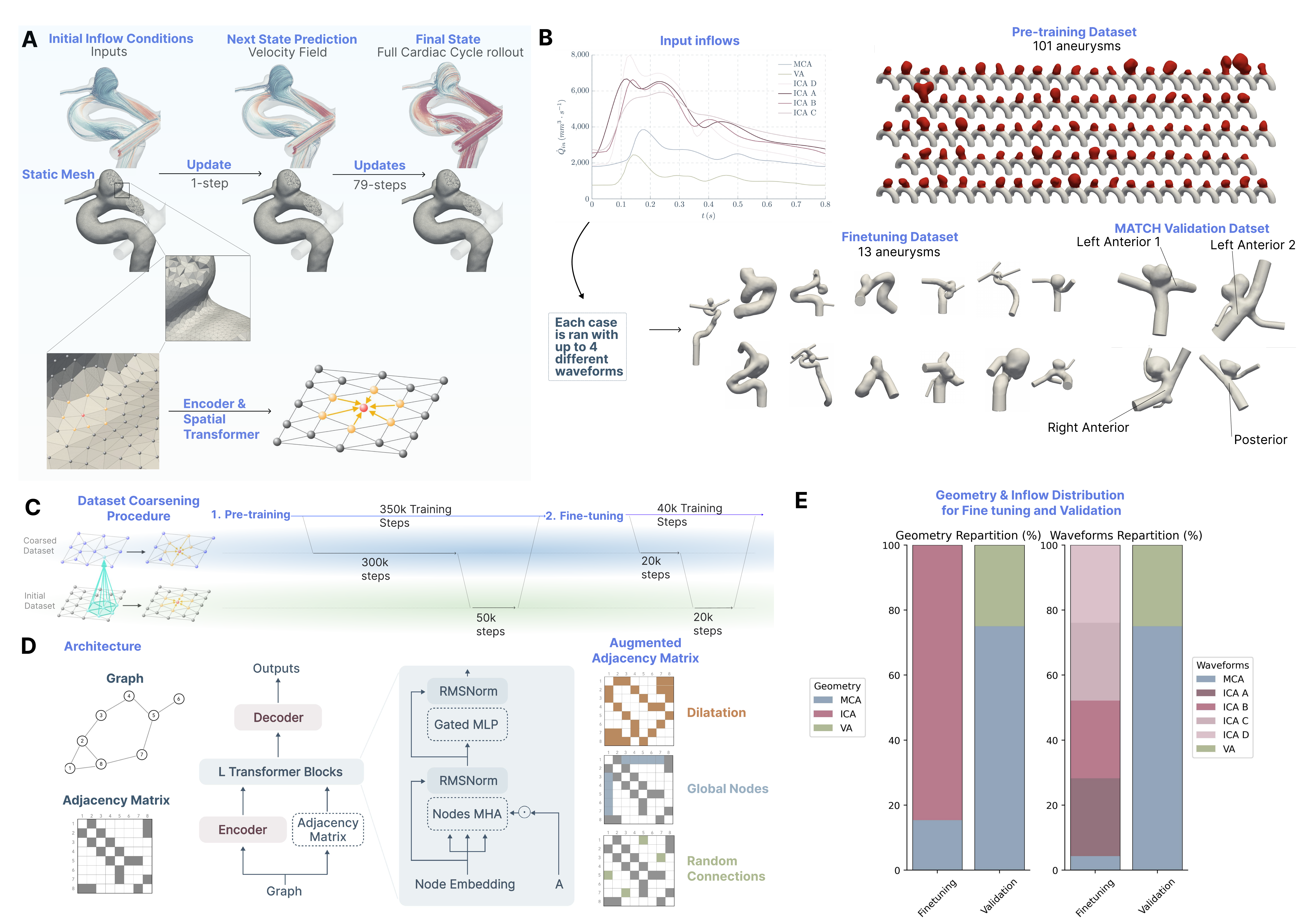}
	\caption{\small
		\textbf{A)} The inputs are defined on an unstructured mesh of, on average, 250,000 nodes. They are comprised of the velocity, the acceleration, the positions and types of the nodes, and finally, statistics about the inflow boundary conditions. In total, this makes for 15 variables per node. The inflow conditions follow the usual values from a cardiac cycle. An example of a mesh with a slice inside the aneurysm bulge is also presented. The model makes a one-step ($\Delta t = 0.01s$) prediction and auto-regressively predicts an entire cardiac cycle ($\approx 0.8s$).
		\textbf{B)} The pre-training dataset comprises 101 standardized aneurysms from the AnXplore dataset. The fine-tuning dataset comprises 13 real aneurysms, simulated with up to 6 different input waveforms. Finally, the validation set is made of 4 geometries (for a total of 5 aneurysms) from the MATCH challenge.
		\textbf{C)} The model is trained in 2 main phases (pre-training and fine-tuning), each split into two sub-phases. The first sub-phase sees the model being trained on a coarse version of the meshes. This allows for much faster training steps. The second sub-phase trains the model on the regular (or fine) meshes. During the first sub-phase, we split the training into 2 (both of 150k steps). In the first split, the coarse meshes are masked to make predictions harder and enhance generalization, similar to a Masked AutoEncoder strategy.
		\textbf{D)} Our model takes the graph nodes' features and the adjacency matrix as inputs. Each transformer comprises masked Multi-Head Self-Attention followed by a Gated MLP with residual connection and Layer Normalization. The Multi-Head Self-Attention masks the node's features by computing a sparse matrix multiplication indexed on the Augmented Adjacency matrix. The adjacency matrix is improved with Dilation, Global Attention, and Random Jumpers.
		\textbf{E)} The distribution of geometries and waveforms between the fine-tuning dataset and the validation set are highly dissimilar. The validation set exhibits highly out-of-distribution patterns, both in terms of geometry and waveforms. This makes for a challenging task in terms of generalization.
	}
	\label{fig:model}
\end{figure}

The full pipeline is illustrated in \autoref{fig:model}, which details the mesh-based input features (\autoref{fig:model}-\textbf{A}), dataset composition (\autoref{fig:model}-\textbf{B},\textbf{E}), multi-stage training strategy (\autoref{fig:model}-\textbf{C}), and the sparse attention model architecture (\autoref{fig:model}-\textbf{D}).

The final model contains 51 million parameters and was trained on approximately 2.85 billion node-time-step pairs across all datasets. Once trained, it can generate a full cardiac cycle prediction on graphs with up to 4 million elements in under one minute using a single GPU\footnote{Using partitioning algorithms such as METIS \cite{Karypis_1998} and multiple GPUs in parallel, we were able to achieve similar simulations per second performances even on larger meshes.}, achieving a $\times$200 speedup over traditional solvers, which typically require hours of computations. We perform extensive analyses and comparisons between the GNN prediction and the ground truth on multiple metrics. We also extract several risk assessment criteria from the MATCH challenge~\cite{Berg2019}. Similar to \cite{kaplan2020scalinglawsneurallanguage} and \cite{hoffmann2022trainingcomputeoptimallargelanguage}, we start by training our model with several set of parameters and for different numbers of training steps in order to extract scaling laws. These scaling laws gives us the optimal amount of training steps and model parameters with respect to the amount of training data we have. Following said laws, the model was trained in five phases described in \autoref{fig:model}-\textbf{C}.

The obtained results demonstrate that our model's predictions remain within 5\% of CFD ground truth across key metrics (velocity, pressure, WSS, OSI). In addition, we benchmark the model's clinical utility by comparing its outputs against the phase diagram risk categories used in hospital practice. Our results suggest that a trained GNN surrogate can replicate and in many cases refine standard risk assessments, while requiring no CFD expertise. This establishes the feasibility of deploying deep-learning-based hemodynamics in clinical settings, replacing hours of simulation and manual intervention with rapid, reproducible inference. Indeed, end-to-end, including segmentation, meshing, and inference, the full prediction pipeline completes in less than 10 minutes. This efficiency opens the door to bedside hemodynamic assessments, from blood flow analysis and risk stratification to virtual planning and surgical intervention optimisation.

\section{Dataset and Simulation}\label{sec:dataset}

\paragraph{Geometries}
Brain arteries, unlike other vascular structures, especially when affected by intracranial aneurysms (IA), are particularly challenging to segment automatically. Vessels at and around the Circle of Willis vary significantly across patients, often featuring hypoplastic or entirely absent segments\,\cite{Hindenes2020}. Moreover, the irregular and non-convex morphology of aneurysms further complicates the use of semi-automatic segmentation tools that work well for larger and more regular structures such as the aorta\cite{Koo2025-jq}. These challenges inherently limit the size and quality of datasets available for model training and evaluation.

To address this, we curated a multi-resolution dataset using subsets of geometries with varying complexity levels and sizes. The first and least complex subset, consists of our database (called AnXplore) gathering 101 Internal Carotid Artery (ICA) aneurysms \cite{Goetz2024b} extracted from their native vascular context \cite{intra} and placed on a toroidal segment emulating the ICA siphon. The second subset is a few-shot training set comprising 13 patient-specific geometries, each retaining an extended segment of the parent vessel and, in most cases, multiple outlets. These cases originate from previous studies \cite{Jeken-Rico2023} and the aforementioned \textit{Intra} dataset \cite{intra}, where the main inclusion criterion was the quality of the lumen surface mesh. In contrast to the AnXplore dataset, this set includes IA not only at the ICA but also at the Middle Cerebral Artery (MCA). To improve the physiological diversity of the training distribution, simulations were conducted under five distinct inflow waveforms \cite{Hoi2011, Najafi2021, Reymond2009}, selected to reflect various hemodynamic states (see \autoref{app:sec:dataset} for additional details).

Finally, model validation was performed using five geometries from the MATCH challenge \cite{Berg2018, voss2019, Berg2019}, including four MCA aneurysms and one on the Vertebral Artery (VA). These cases were thoroughly curated in the original publication and serve as an out-of-distribution benchmark, statistically distinct from the training data, as illustrated in \autoref{fig:res2}-\textbf{A} and \autoref{fig:model}-\textbf{E}.

Each surface mesh was converted into a high-resolution volumetric mesh composed solely of tetrahedral elements. Mesh sizes ranged from 0.1 to 0.2mm, in line with standard practices in literature,\cite{Ahn2014, Berg2019rev}. To accurately resolve near-wall shear gradients, we added structured boundary layers with an initial thickness of 20$\mu$m and a geometric growth factor of 1.2.

In total, the dataset comprises 147 high-resolution simulations based on meshes averaging 250k nodes used for model training and validation, with an additional 5 simulations reserved exclusively for final testing.

\paragraph{Hemodynamics Simulations}
Blood flow within the cerebral arteries and aneurysms was simulated by solving the incompressible Navier–Stokes equations for a single-phase fluid with a density of 1056kg/m$^3$ \cite{Ahn2014}. Recognizing that blood exhibits non-Newtonian behavior at physiologically relevant scales due to red blood cell interactions, we modeled its rheology using the shear-thinning modified Casson model \cite{Suzuki2021}. This choice is particularly important, as shear-thinning effects, driven by erythrocyte alignment, may influence slow or recirculating flow patterns that are often observed within aneurysms \cite{Nader2019}. A standard hematocrit (Hct) value of 40\% was assumed for the simulations.

The governing equations were solved using an in-house finite element solver \cite{digonnet2007cimlib} based on linear elements. To ensure numerical stability, particularly in convection-dominated regimes, and to satisfy the compatibility conditions between velocity and pressure interpolations, a Variational Multiscale stabilization technique was adapted for blood flow \cite{Hachem2010}. Time integration was performed using a second-order, semi-implicit scheme with a timestep of 1ms, selected to maintain Courant–Friedrichs–Lewy numbers within a stable range. Simulations were run for at least two full cardiac cycles to eliminate initial transients, with only the final cycle retained for analysis.

Boundary conditions were defined for inlets, outlets, and vessel walls. Patient-specific inflow waveforms, derived from cine phase-contrast magnetic resonance imaging, were applied at the inlets, corresponding to the parent vessel, either the ICA \cite{Hoi2011, Najafi2021}, MCA \cite{Reymond2009}, or VA \cite{Reymond2009}. To account for physiological variability, four distinct ICA waveforms reflecting different demographic profiles were used. These flow rates were converted into transient parabolic velocity profiles at the inlets. Outlets were modeled using traction-free conditions, and vessel walls were assumed to be rigid with a no-slip boundary condition, consistent with established CFD practice \cite{Berg2019rev}. Finally, cylindrical extensions were added to the inlet and outlet boundaries to reduce the influence of the imposed boundary conditions on the flow field \cite{Berg2019rev, Jeken-Rico2023}.

Analysis focused on key hemodynamic indicators that are strongly associated with aneurysm pathophysiology. WSS, which represents the tangential force of blood flow exerted on the vessel wall, plays a central role as it directly influences endothelial cell function and arterial wall remodeling. Deviations from physiological WSS levels have been implicated in aneurysm initiation, progression, and rupture risk \cite{Urschel2021, Zimny2021, Meng2014}. Our analysis primarily targeted the Time-Averaged Wall Shear Stress (TAWSS), which highlights regions of persistently high or low shear. Pathologically low TAWSS is associated with pro-inflammatory states, endothelial dysfunction, and progressive wall weakening. Conversely, some studies have shown that abnormally high WSS may also promote rupture via distinct mechanisms, such as inducing local hypocellularity and tissue degradation \cite{Cebral2017}.

In addition to TAWSS, we computed the OSI, which quantifies the directional changes in WSS over the cardiac cycle \cite{Zhou2017, Xiang2014}. Elevated OSI indicates disturbed, non-unidirectional flow, a known contributor to endothelial injury, atherogenesis, and wall remodeling leading to rupture. Finally, we qualitatively assessed the overall flow dynamics, particularly the presence of recirculation zones, secondary vortices, and instabilities, as such features have also been linked to increased rupture risk in several studies \cite{Li2019, Chung2018}. For a more detailed explanation of the hemodynamic indicators used, we refer the reader to \autoref{app:sec:dataset}.

\section{Graph Neural Network}\label{sec:model}

We introduce the transformer-based GNN, a 51M parameters model with an augmented adjacency matrix. Trained on one-step predictions, it generates autoregressively a complete cardiac cycle on meshes with up to 4M elements from a single initial state. A single cardiac cycle takes 1 minute on a single A100 GPU, while the simulation using classical CFD methods requires several hours with 64 CPU cores. 
The proposed model takes a graph $G_t$ as input and predicts $G_{t+1}$ (see \autoref{fig:model}-\textbf{A}). The graph only holds per node features: the current velocity and acceleration, the node's position, its type and distance to inflow, and finally, statistics regarding the inflow velocity (see \autoref{appendix:inputs} for details). Each time step is separated by $\Delta _t = 0.01s$. We enforce non-slip boundary conditions on the vessels and aneurysm walls by setting the velocity field to zero.

The model follows an Encode-Process-Decode architecture with a Transformer processor, following the works from \cite{sanchezgonzalez2020learning,pfaff2021learning,garnier2024transformer}. The architecture works by first encoding the node features in a latent space $\mathbf{Z} = (\mathbf{z}_1, \mathbf{z}_2,...,\mathbf{z}_N)^{T} \in \mathbb{R}^{N \times d} = \mathcal{E}(G)$. We then process $\mathbf{Z}$ through $L$ Transformer Blocks (see \autoref{fig:model}-\textbf{D}). Each block takes the latest latent representation $\mathbf{Z}$ and the Adjacency Matrix $\mathbf{A}$ as input:

\begin{align}
    \mathbf{Z'}_l &= \text{RMSNorm}\big(\text{MMHA}(\mathbf{Z}_{l-1}, \mathbf{A}) + \mathbf{Z}_{l-1}\big) && \ell \in [1\ldots L] \label{eq:msa_apply} \\
    \mathbf{Z}_l &= \text{RMSNorm}\big(\text{GatedMLP}(\mathbf{Z'}_{l}) + \mathbf{Z'}_{l}\big) && \ell \in [1\ldots L] \label{eq:mlp_apply}
\end{align}

$\text{MMHA}$ is a Multi-Head attention operation masked by the Adjacency matrix:

\begin{equation}
    \text{MMHA}(\mathbf{Z}) = \Bigg(\mathbf{A}\odot\sigma\Big( \frac{QK^T}{\sqrt{d}} \Big)\Bigg)V
\end{equation}

where $Q, K, V$ are learned linear projections of $\mathbf{Z}$.

$\text{GatedMLP}$ is a gated Multi-layer perceptron following the work of \cite{dauphin2017language}:

\begin{equation}
    \text{GatedMLP}(\mathbf{Z}) = W_f\Big(\text{GeLU}\big(W_l \mathbf{Z} + b_l\big) \odot (W_r \mathbf{Z} + b_r)\Big) + b_f
\end{equation}

and RMSNorm is a normalization layer \cite{zhang2019rootmeansquarelayer}.

Finally, a decoder maps back $\mathbf{Z}_L$ into a physical space $\mathbb{R}^3$ to predict the blood flow's velocity at the following time step. Both the encoder and the decoder are 2-layers Multi-layer perceptron, where the second layer's width (\textit{resp.} the first layer for the decoder) is set to $d$. They also use an RMSNorm layer and a $\text{ReLU}$ activation function.

Similar to the Longformer and BigBird \cite{beltagy2020longformerlongdocumenttransformer, zaheer2021bigbirdtransformerslonger}, the adjacency matrix is augmented using random connections to increase the model's receptive fields, global attention to propagate the variable inflow to each node, and 2-Dilation to propagate information further away at each processor block (see \autoref{fig:model}-\textbf{D}). The random connections add edges between pairs of disconnected nodes, allowing information to travel much longer distances. The 2-Dilation uses the standard adjacency matrix $\mathbf{A}$ for half of the heads of each processor block and $\mathbf{A}^2$ for the remaining heads. This lets information propagates with receptive fields twice as big at every layer.

During training and inference, inputs and outputs are normalized to zero-mean and unit variance. We use a Mean Squared Error (MSE) averaged on each node as a training loss. Following the work of \cite{sanchez2020}, we add a multivariate gaussian noise centered on zero to some inputs such as the velocity to make our model robust to error propagation over multiple time steps\footnote{For the cases from AnXplore, since the vessels are always oriented in the same direction (mostly in the $x$ and $y$ axis, we add less noise in the $z$ axis, using $\mathbf{\sigma} = [10,10,1]$. For the other cases, we use $\mathbf{\sigma} = [10,10,10]$. All values are in mm/s}.

The model was trained in 5 phases (see \autoref{fig:model}-\textbf{C}). We first train it using masking on the coarse pre-training dataset, following the strategy from \cite{garnier2025meshmask}, followed by a second phase without masking on the same dataset. We then train it on the regular pre-training dataset. Finally, we fine-tune it on a coarse and then fine version of the second dataset. Overall, all training is split into a coarse then fine sets, following works on curriculum learning \cite{bengio,garnier2025curriculumlearningmeshbasedsimulations}. While we perform the final validation on a specific dataset from the MATCH challenge, we also keep 10\% of each dataset for testing during training. More details are available in \autoref{app:sec:training}. 

Overall, we trained the model for 430k gradient descents with a batch of size 2, translating into roughly $3.5 \times 10^{7}$ training PFLOPs\footnote{Peta FLOPs}. We also trained up to 60 variants allowing the model to discover a scaling law between the number of FLOPs and the number of parameters.

\section{Results and Discussion}\label{sec:results}

We begin by benchmarking our method against leading mesh-based machine learning models\footnote{We also trained and evaluated a Transolver \cite{luo2025transolverpp} model but didn't get good enough performances.}, including MeshGraphNet \cite{pfaff2021learning}, BSMS-GNN \cite{cao2023efficientlearningmeshbasedphysical}, and Attention-Multigrid \cite{garnier2025multi}. To ensure a fair comparison, we follow the exact training protocols described in their respective reference papers. All models are trained for the same number of gradient steps, using identical optimizers, learning rate schedules, and tuned hyperparameters (such as width and depth). We also verify that each model is trained to convergence, and evaluations are conducted on a shared test suite under consistent conditions.

To explore the optimal capacity of our transformer-based architecture, we construct an empirical scaling law. Specifically, we model the final All-Rollout RMSE on the validation set as a function of $P$, the number of trainable parameters, and $D$, the number of training nodes. For six different FLOPs budgets (see \autoref{fig:scalinglaws}), we vary model width and depth to evaluate performance across a wide range of configurations. The resulting relationship follows $P \propto C^{0.75}$, where $C$ is the computational budget in FLOPs. Guided by this law and the characteristics of our datasets, we select a final model with $L = 15$ layers and hidden width $d = 512$, totaling 51M parameters.

We also perform an ablation study to assess the trade-off between model size and training efficiency. Despite having an order of magnitude more parameters than some baselines (see \autoref{fig:overview}), our model exhibits similar training speed and memory footprint, owing to its optimized transformer structure and sparse attention formulation.

On both validation datasets, the in-distribution AnXplore set and the out-of-distribution MATCH challenge set, our model outperforms all prior baselines by a wide margin, reducing error by nearly 2$\times$ relative to the strongest GNN competitor (see \autoref{fig:res1}-\textbf{C} and \autoref{fig:res2}-\textbf{D}). During pretraining, we compare against MeshGraphNet, BSMS-GNN, and Masked Multigrid; for the final validation phase, we retain MeshGraphNet and Masked Multigrid as strong references.

On the AnXplore dataset, our model reproduces CFD solutions with a mean accuracy of up to 99\%. Running our model autoregressively on the 10 held out geometries (\emph{i.e.} 790 total timesteps), we obtain an averaged $L_2$ error of $2.8 \pm 0.7$ mm/s inside the aneurysm's bulge. 

On the more challenging MATCH validation set, featuring out-of-distribution geometries and inflow waveforms, accuracy remains above 93\%. Notably, one validation case includes two aneurysms within the same parent vessel, significantly increasing flow bifurcation complexity and prediction difficulty (see \autoref{fig:res2}-\textbf{A}). Running our model on the 5 aneurysms, we obtain an averaged $L_2$ error of $5.4 \pm 1.1$ mm/s inside the aneurysm's bulge.  

The predictive speed-up is substantial. While high-resolution CFD simulations typically require 3-4 hours per case, our trained model delivers full-cycle predictions in under 1 minute. Even accounting for dataset generation and training time (approximately 20 days total), the break-even point is reached after inference on just 100 aneurysm cases; making the method highly practical for large-scale deployment (see \autoref{fig:datasetsdetails-time}). This speed-up can be even better if we start to work with simulations that uses endovascular devices, or under a Fluid Structure Interaction paradigm.

In the following sections, we compare the high-fidelity CFD simulations and our transformer-based GNN across multiple metrics. These include the full velocity field (predicted autoregressively; see \autoref{fig:res1}-\textbf{A}), Wall Shear Stress (WSS), Time-Averaged WSS (TAWSS), and Oscillatory Shear Index (OSI). The relative change between the ML prediction and the CFD reference is quantified as:

\begin{align}
	\Delta \text{metric} [\%] = 100 \times \frac{\text{metric}_{\text{GNN}} - \text{metric}_{\text{CFD}}}{\text{metric}_{\text{CFD}}}
	\label{eq:delta_metric}
\end{align}

\subsection{Flow analysis}

\begin{figure}[!ht]
	\centering
	\includegraphics[width=1\textwidth]{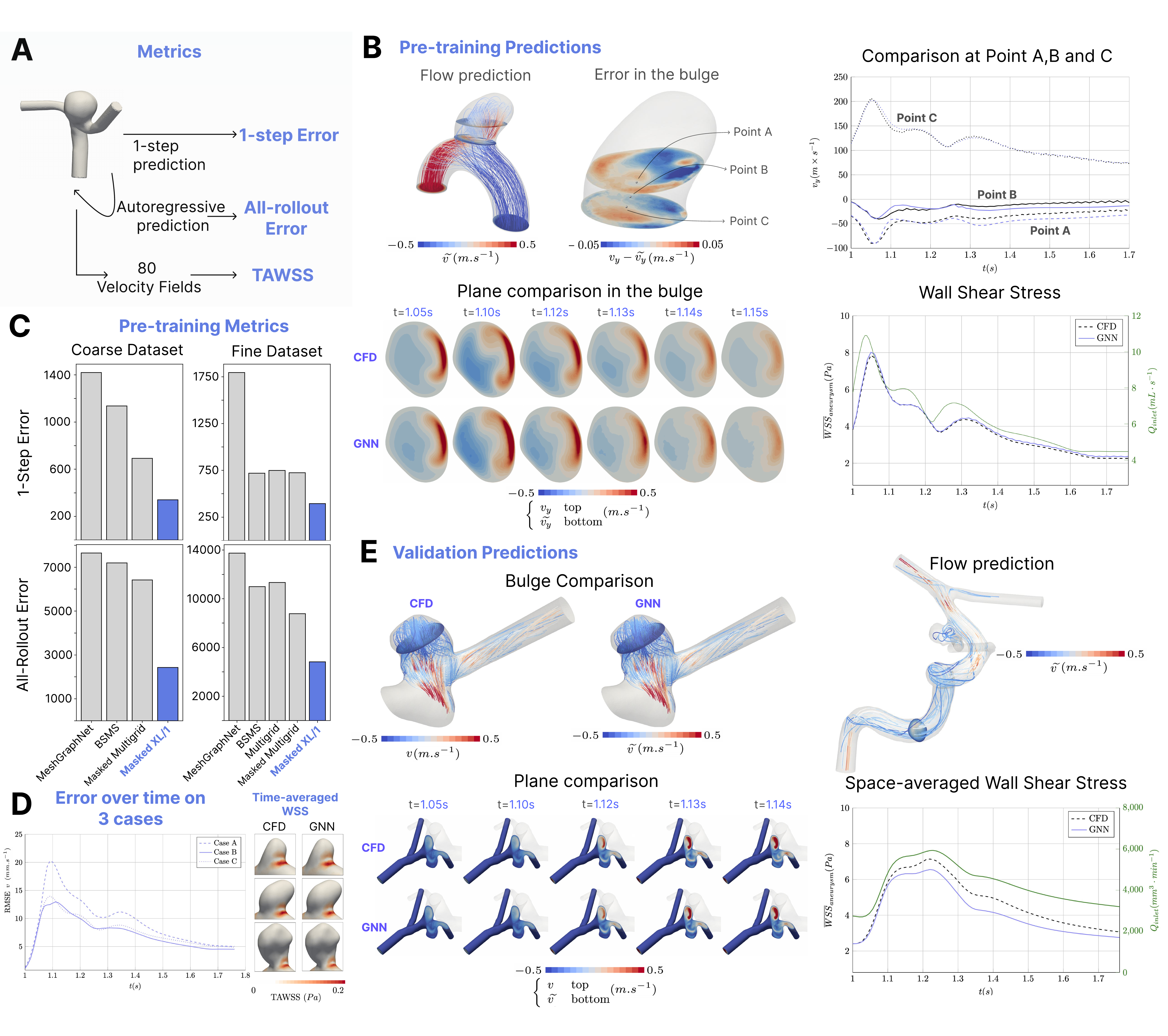}
	\caption{\small
		\textbf{A)} Our model makes a one-step prediction by default. We use this to compute a 1-step error metric. We also use our model in an autoregressive fashion to predict an entire cardiac cycle and compute an All-rollout error based on all time steps from the cardiac cycle.
		\textbf{B)} We showcase results on the pre-training dataset. We display predicted systolic flow, errors inside the bulge at $y=8$ and $y=10$, a comparison of three selected points at $y=8, y=9, y=10$, and a comparison between the CFD (top row) and our transformer (bottom row) for $v_y$ inside the bulge in a 2D plan defined by $x=0, y=10, z=0$. The GNN model accurately captures both velocity fields and derived wall shear features. Additional visualisation can be found in \autoref{fig:res-pretraining}.
		\textbf{C)} Our model outperforms every physical-based GNN on the pre-training dataset on 1-step and All-rollout errors. Results are $\times 10^{-3}$.
		\textbf{D)} Similar to subfigure B), our GNN follows very closely the ground truth (less than 1\% error) of the TAWSS on a full cardiac cycle. The RMSE is in mm/s. 
		\textbf{E)} We showcase results on the validation. We display predicted systolic flow in the artery and the bulge and a plane comparison inside the shapes between the CFD (top row) and our transformer (bottom row). Overall, flow directions are very well predicted, but our model often lacks velocity magnitude (around 10\% on average). Additional visualisation can be found in \autoref{fig:res-validation}.
	}
	\label{fig:res1}
\end{figure}

\autoref{fig:res1} illustrates the performance of our GNN-based approach across several representative cases, focusing primarily on the velocity field and derived WSS. Velocity serves as a foundational variable in hemodynamic modeling, as it directly informs the computation of shear forces and is itself an established marker of aneurysm complexity and rupture risk\,\cite{Chung2018, Li2019}. The interplay between aneurysm morphology, parent vessel geometry, and inflow waveform heavily influences the resulting flow patterns. As such, both the computational demands of traditional CFD and the predictive accuracy of learning-based models are sensitive to these anatomical and physiological variations. Our model consistently captures the spatiotemporal dynamics of the flow, including subtle near-wall effects, while dramatically reducing simulation time and maintaining high fidelity across training and validation cohorts.

\paragraph{AnXplore Validation Set}
\autoref{fig:res1}-\textbf{B} presents the results of our model on a randomly selected test cases from the pre-training (AnXplore) dataset (additional results can be found in \autoref{fig:res-pretraining}). The plane cuts show that velocity magnitude errors remain below 5\% within the core flow regions, indicating high fidelity. The $y$-component of the velocity exhibits the largest relative deviation, which, for the cases shown, coincides with localized numerical instabilities observed in the CFD baseline. These discrepancies are likely artifacts of the trajectory sampling timestep\footnote{The timestep referenced here denotes the frequency at which velocity fields were recorded; CFD simulations used an internal timestep of $10^{-3}$\,s.}, and notably, they are not reproduced by the GNN, underscoring its robustness. Temporal snapshots of the $v_y$ component across several cut planes further demonstrate strong qualitative agreement between the model and the ground truth. Notably, even at timesteps that reflect very different flow dynamics such as the systole and then the diastole, our model retain a strong variance and is much more than simply an inflow average operator.

Finally, the rightmost plots compare the inflow waveform with the bulge-averaged WSS, revealing tight correspondence and minimal overestimation by the GNN. This confirms the model’s capacity to faithfully capture both bulk dynamics and wall-level shear behavior across physiologically relevant flow regimes.

\begin{figure}[!ht]
	\centering
	\includegraphics[width=1\textwidth]{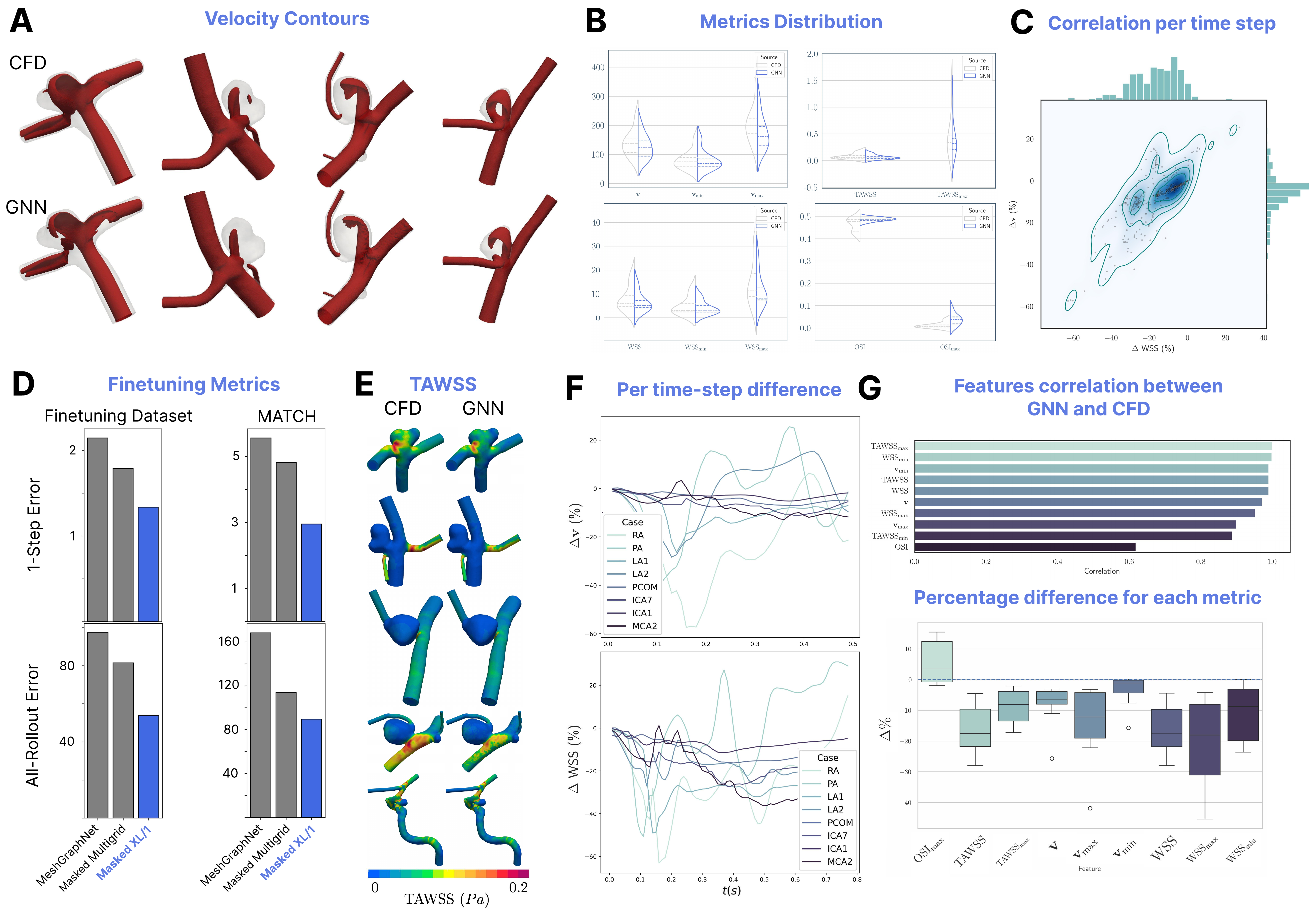}
	\caption{\small
		\textbf{A)} Velocity contours at systole for the validation cases. The chosen velocities are, from left to right, 600, 100, 200, and 100\,mm/s.
		\textbf{B)} Distribution between CFD and GNN on the validation and test sets from the fine-tuning. We compare them for $\mathbf{v}$, $\mathbf{v}_{max}$, $\mathbf{v}_{min}$, $\text{WSS}$, $\text{WSS}_{max}$, $\text{WSS}_{min}$, $\text{TAWSS}$, $\text{TAWSS}_{max}$, $\text{OSI}$ and $\text{OSI}_{max}$.
		\textbf{C)} Evolution on the velocity and WSS (following \autoref{eq:delta_metric}) per time step for all validation cases, plotted as a density map.
		\textbf{D)} Performance of our model on the fine-tuning dataset and the validation set against two other GNNs approaches. Results are $\times 10^{-3}$.
		\textbf{E)} TAWSS distribution on the four validation cases and 1 case from the fine-tuning test set.
		\textbf{F)} Evolution of the velocity and WSS per time step for all validation cases.
		\textbf{G)} Correlation coefficient between CFD and GNN for several important metrics. We also display the evolution of said metrics as relative changes between CFD and GNN, following \autoref{eq:delta_metric}
	}
	\label{fig:res2}
\end{figure}

\paragraph{MATCH Validation set}
The validation set displayed in \autoref{fig:res1}-\textbf{E} and \autoref{fig:res2} incorporates geometrical features, such as branches at the aneurysm and larger bulge sizes, which increase the flow complexity. Despite this and the use of other waveforms, the GNN manages to provide strong estimations of the flow structures in both the parent vessels and the aneurysms, as seen in \autoref{fig:res1}-\textbf{E} and \autoref{fig:res2}-\textbf{A} (additional results can be found in \autoref{fig:res-validation}). One observed issue remains the magnitude of the flow velocity, which is underestimated by 10\% in some cases. This possibly results from evaluating the model on MCA and VA aneurysms with inflow rates that differ considerably from those used in the training cases. Indeed, on a held out geometry from the validation set, we see no such underestimation. The WSS, which is positively correlated with the velocity (see \autoref{fig:res2}-\textbf{B}, \textbf{C}), exhibits similar behavior with dropping mean WSS values. It qualitatively agrees with the CFD results in all, but two validation cases. Looking at the TAWSS (\autoref{fig:res2}-\textbf{E}), one can see a strong qualitative agreement in both vessel and aneurysm regions. Peak values, such as those found at bifurcations or vessel turns, concentrate most of the underestimation errors.

Overall, our model can generate accurate velocity fields, even on intricate shapes that lie far outside the training distribution in terms of geometry and input waveforms, with only a slight downside in terms of pure magnitude. As an experiment, we closed one of the artery outflow boundaries and predicted the flow during a cardiac cycle with our model. The model accurately predicts the blood flow rebound on the wall and return to other outlets, proving its strong generalization capabilities.

\subsection{Risk metric analysis}
Analysis of the hemodynamic risk metrics on the validation set revealed distinct patterns. Based on the averaged TAWSS over the aneurysm bulge, the five aneurysms were separated into two groups. A single aneurysm (left-most in \autoref{fig:res2}-\textbf{A}) exhibited a high mean TAWSS of 3.34 Pa, with the GNN model predicting a value differing by 1.17\,Pa. This level of TAWSS is characteristic of locations like the M1 MCA terminus\,\cite{Zhou2017}. The remaining four aneurysms presented a significantly lower average TAWSS of 0.29 Pa, with a proportionally similar mean prediction error of 0.11\,Pa. According to thresholds proposed by Zhou et al.\,\cite{Zhou2017}, this lower TAWSS range might indicate a potential risk zone. However, classifying aneurysm risk based solely on these thresholds remains challenging. Reported values in longitudinal studies often exhibit large variability, with standard deviations frequently comparable to the mean values themselves, complicating reliable classification\,\cite{Chung2018, Zhou2017, Lee2021}.

While the GNN model resolved the core flow features and overall TAWSS magnitudes reasonably well, we encountered lower accuracy regarding the OSI. Difficulties arose primarily in regions of low TAWSS where the GNN model predicted slight variations or intermittencies in the near-wall flow direction. The OSI calculation is known to be highly sensitive to such fluctuations in the WSS vector direction. These limitations in predicting complex near-wall flow markers like OSI might stem from the model's training objective, which prioritized matching overall flow profiles rather than specifically optimizing for boundary layer accuracy. 

The last, most recent risk indicator that we consider is the complexity of the flow. The indicator was particularly acclaimed in the context of larger aneurysms which would rupture overproportionally when secondary and more complex flows were present in the bulge\,\cite{Li2019, Chung2018}. In the present validation cases, the GNN model manages to predict the bulk without sizable qualitative errors apart from the mentioned reduced velocity magnitude.

\subsection{Clinical validation and risk stratification}

To bridge the gap between simulation fidelity and clinical utility, we evaluated the five validation cases using both the established PHASES score \cite{Greving2014} and our GNN-based hemodynamic predictions. The PHASES score is routinely used in clinical settings to stratify unruptured aneurysms according to their estimated 5-year rupture risk. Independently, we computed a hemodynamic risk score using thresholds for systolic velocity, WSS, TAWSS, and OSI, based on the classification system proposed in previous studies \cite{Zhou2017, Chung2018, Sheikh2020, Axier2022, Wei2022, Xin2022_CombinationMorphHemodyn, Tian2022}. We compute an average of 4 risks: systolic velocity [mm/s] ($<$20=1, 20-50=0, 50-80=1,$>$80=2), peak WSS [Pa] ($<$4=0, 4-6=1,$>$6=2), OSI ($<$0.15=0,0.15-0.30=1,$>$0.3=2) and TAWSS [Pa] ($<$1.5=1,1.5-6.7=0,$>$6.7=2) (see \autoref{appendix:risk} for more details). 

While no scientific consensus exists in regards to a strong IA rupture metric, we conducted an extensive study to deeply root this metric in the literature. However, we do not run any large scale statistical analysis and study its validity, we only define the metric above mostly as a practical additional metric to compare the CFD and the GNN aproaches. 

Table~\ref{tab:clinical-validation} summarizes the rupture risk classification for each aneurysm case. PHASES scores were generally low, consistent with conservative treatment strategies. In contrast, our GNN model occasionally detected elevated hemodynamic risk patterns, often in agreement with ground truth CFD values. The agreement between our predicted hemodynamic risk and the CFD-derived baseline was high: in 3 out of 5 cases, the GNN and CFD-derived average risk score differed by at most 0.25 points, the 2 others achieving the same results.

\begin{table}[htbp]
\centering
\caption{Comparison between clinical PHASES score, CFD results and GNN-predicted hemodynamic risk. Scores are classified as Low ($<$1), Moderate ($\in[1,2)$), or High ($\geq 2$).}
\label{tab:clinical-validation}
\begin{tabular}{lcccc}
\toprule
\textbf{Case} & \textbf{PHASES Score} & \textbf{5-yr Risk} & \textbf{CFD Hemodynamic Score} & \textbf{GNN Hemodynamic Score} \\
\midrule
LA1 & 3 (Low) & 0.7\% & 1.00 (Moderate) & 1.25 (Moderate) \\
LA2 & 3 (Low) & 0.7\% & 0.75 (Low) & 1.00 (Moderate) \\
P    & 5 (Moderate) & 1.3\% & 0.50 (Low) & 0.50 (Low) \\
Pcom & 5 (Moderate) & 1.3\% & 0.75 (Low) & 0.75 (Low) \\
RA   & 3 (Low) & 0.7\% & 0.75 (Low) & 1.00 (Moderate) \\
\bottomrule
\end{tabular}
\end{table}

These results demonstrate that our GNN model can infer not only velocity and WSS fields, but also approximate complex clinical risk markers such as flow instability and abnormal shear environments. In particular, the elevated scores for LA1 and RA suggest that our model is capable of revealing latent hemodynamic risks not captured by demographic-based stratification alone. Such insights could be clinically valuable, especially for borderline or ambiguous cases where traditional morphological criteria fall short.

Notably, even in cases where the PHASES score predicted low rupture risk, the GNN model aligned well with CFD estimates in identifying moderate hemodynamic stress environments. This reinforces the idea that AI-based models, when trained on high-fidelity CFD data, can extend beyond mere acceleration of simulation and act as clinical adjuncts for risk-aware treatment planning.

Taken together, this hybrid AI–clinical stratification pipeline offers a compelling pathway toward real-time, explainable, and robust decision support in cerebrovascular care.

\section{Conclusion}\label{sec:conclusion}

This work introduced a high-fidelity, autoregressive GNN with transformer architecture capable of accurately predicting blood flow fields, shear stresses, and derived hemodynamic risk indicators in cerebral aneurysms, with over a 200 fold speed-up compared to traditional high fidelity computational methods. It's important to note that for more complex cases, such as Fluid-Structure Interaction simulations or stented ones, the speed-up could be an order of magnitude greater. Our model, trained on synthetic cases and validated on challenging out-of-distribution geometries and waveforms, achieved up to 99\% agreement with CFD baselines on the AnXplore dataset and maintains robust performance (93\%) on unseen clinical cases, with errors ranging around 2.5 and 5 mm/s.

Through a systematic scaling law, we identified the optimal architecture size (51M parameters) balancing expressiveness and training efficiency. Despite its relatively modest size, the model demonstrated strong generalization and inference capabilities, and remained trainable on standard hardware. By enforcing velocity boundary conditions and using only the predicted velocity fields as input, our method approximated downstream hemodynamic quantities such as WSS, TAWSS, and OSI with remarkable fidelity, even though these fields were never directly supervised during training.

Importantly, we evaluated the clinical relevance of our predictions by comparing GNN-derived hemodynamic risk scores against both ground truth CFD and the PHASES score, a widely used clinical tool. In all validation cases, our GNN model matched CFD-based risk classifications within 0.25 points, confirming its reliability. More notably, in cases where the PHASES score underestimated risk, our model identified latent hemodynamic abnormalities, offering a promising complementary signal for surgical decision-making. These findings pave the way for hybrid stratification pipelines that couple morphological scores with fast, data-driven flow predictors, enabling real-time and explainable treatment planning. It's still important to note a strong limitations of both those risk metrics: we merely use them as a way to compare our approach with a standard CFD solver, and not as perfect aneurysm rupture prediction score.

Our approach offers a scalable foundation for a new class of clinical support tools. Reducing computation times from hours to minutes makes large-scale simulation campaigns, such as population-wide risk stratification, practical within a clinical timeframe. Several improvements are envisaged, in particular to explore the integration of physics-informed loss functions, boundary-aware learning, and multi-task supervision of shear-based metrics to further enhance accuracy, especially in near-wall regions. Adding fast, steady-state features could also help mitigate gradient-related issues. Another limitation is the mesh size we currently handle (up to 4M elements). While we experimented with much larger meshes using partitioning techniques, a rigorous scaling analysis is needed to scale to meshes with tens of millions of elements.

Future work will integrate the presence and effects of intracranial stents, allowing the model to simulate post-intervention flow conditions. This evolution transforms the system from a purely diagnostic tool into a decision-support framework for treatment planning. By enabling rapid assessment of multiple stent configurations per patient, our model holds promise for assisting clinicians in selecting optimal interventions, tailored to each aneurysm’s geometry and flow dynamics. 

Ultimately, we envision our model as a key enabler of personalized cerebrovascular care, turning high-fidelity blood flow simulation into an accessible, fast, and interpretable component of clinical workflows.

\newpage
\backmatter

\bmhead{Supplementary information}

Supplementary materials are provided after the references.

\bmhead{Acknowledgements}

The authors acknowledge the financial support from ERC grant no 2021-CoG-101045042, CURE. Views and opinions expressed are however those of the author(s) only and do not necessarily reflect those of the European Union or the European Research Council. Neither the European Union nor the granting authority can be held responsible for them.

\bibliography{main}

\newpage
\begin{appendices}
\tableofcontents
\newpage

\section{Dataset}\label{app:sec:dataset}
\subsection{Shapes}

To meet the demands on both quantity of training data and its physiological accuracy, the dataset is structured in three blocks of ascending complexity and descending number of cases. The fundamental geometrical differences here relate to the inclusion of the vascular context, the quality of the segmentations and the size of the simulated domain.

\subsubsection*{Training set}
\label{subsec:training-set}
\begin{figure}[!ht]
    \centering
    \includegraphics[width=\linewidth]{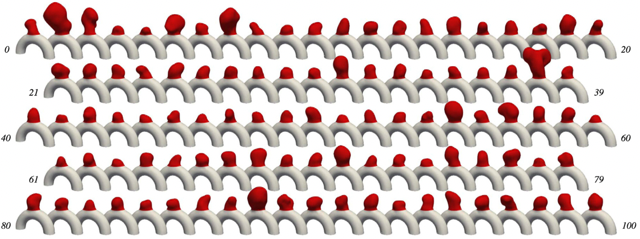}
    \caption{Cases from the AnxPlore dataset.}
    \label{fig:anxplore}
\end{figure}

This simplified set of geometries was obtained from the data published in \cite{Goetz2024a, Goetz2024b}. The cases are generated by scaling and merging segmented aneurysms from the Intra challenge dataset \cite{intra} with a torus ($r$\,=\,2\,mm, $R$\,=\,6.24\,mm) resembling an Internal Carotid Artery (ICA) siphon. These geometries retain the shape irregularities at the bulge, while homogenizing the vascular context, permitting a seamless setup of a larger quantity of cases ($n$\,=\,101) of this type (see \autoref{fig:anxplore}).

The surface mesh is extruded inwards to form a structured boundary layer mesh of a total thickness of 0.3\,mm, a minimal size 0.02\,mm and a commonly chosen progression factor of 1.2. From the last layer onwards, the interior is meshed with isotropic tetrahedra with an edge size of 0.17\,mm. The numbers of elements per mesh are available in \autoref{table:num_elements_dataset}. It would be interesting to improve this dataset with latent visual features from models such as TRELLIS \cite{xiang2024structured}, following the work of \cite{hervé2025trellisenhancedsurfacefeaturescomprehensive}. 

\subsubsection*{Few-shot training set}
\label{subsec:fewshot}
\begin{figure}[!ht]
    \centering
    \includegraphics[width=\linewidth]{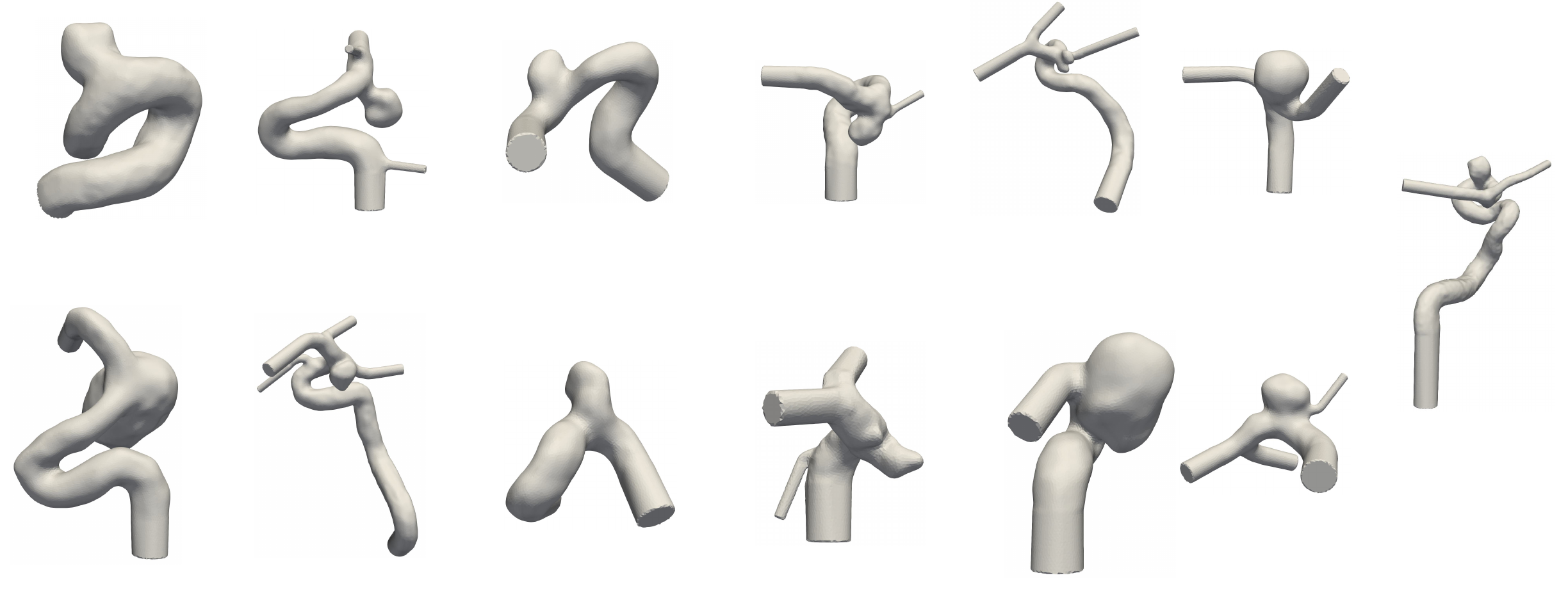}
    \caption{The 13 cases from the Few-shot dataset.}
    \label{fig:fewshot}
\end{figure}

The lumen of further 13 patients including 11 ICA and 2 MCA aneurysms has been processed to provide more physiologically accurate simulations. The segmentations stem from previous works (n=5) \cite{Jeken-Rico2023, Hachem2023} and partly from the aforementioned Intra dataset (n=10)\,\cite{intra}. Contrary to the simplified set, the vascular context is kept intact for these aneurysms including larger portions of the parent vessel and bifurcating branches. Additionally, inlet and outlets are extruded outwards by at least 5\,mm to minimize the impact of the in- and outflow modelling (see \autoref{fig:fewshot}).

To improve the generalizability, these cases were meshed with tetrahaedra with sizes varying between 0.1 and 0.2\,mm and the aforementioned boundary layers. This resulted in mesh sizes ranging from 1M to 3.4M elements.

\subsubsection*{Validation Set}
\begin{figure}[!ht]
    \centering
    \includegraphics[width=\linewidth]{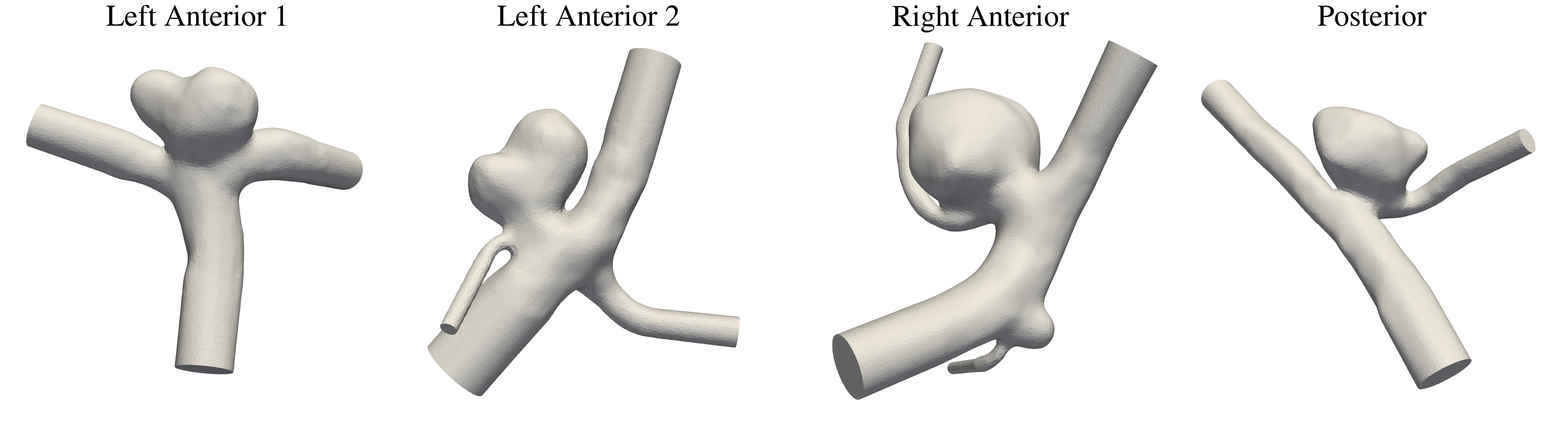}
    \caption{Aneurysm geometries from the MATCH challenge\,\cite{voss2019, Berg2019} used in the final evaluation of the model. The aneurysms found on the anterior circulation are situated on the MCA. The right-most aneurysm originates from the left VA as described in Berg \textit{et al.}\,\cite{Berg2019}.}
    \label{fig:val_aneu}
\end{figure}

The final evaluation of the model was conducted using the well-known MATCH challenge\,\cite{voss2019, Berg2019} in brain aneurysm research. This challenge features five aneurysms (4–6\,mm) located in the anterior and posterior circulations of a single patient. The challenge carefully assessed segmentation results from participating teams by comparing them with high-resolution 7T MRI images. The lumen segmentation from the winning team was utilized to extract four vessel sections, shown in \autoref{fig:val_aneu}. These sections are approximately the size of regions of interest typically analyzed by clinicians using software such as AneurysmFlow (Philips, Best, the Netherlands) or Sim\&Cure (Montpellier, France).

The validation geometries include features absent from the training dataset to evaluate the model’s performance on out-of-distribution cases. While the formations in the left anterior circulation resemble those seen during training (two left images in \autoref{fig:val_aneu}), the right-side examples include two aneurysms with side branches. Finally, the posterior circulation aneurysm (rightmost image in \autoref{fig:val_aneu}) represents a brain location not included in the training set and incorporates a unique inflow waveform.

\begin{table}[h!]
    \centering
    \renewcommand{\arraystretch}{1.5} 
    \setlength{\tabcolsep}{8pt}      
    \begin{tabular}{crlrr}
        \hline
        Dataset  & \# samples & Mesh size & \#Elements \\
        \hline
        AnXplore & 101 & 0.17\,mm & 0.8-1.6M \\
        Few-Shot & 13 & 0.1-0.2\,mm & 1.0-3.4M \\
        Validation & 4 & 0.1\,mm & 1.8-2.8M \\\hline
    \end{tabular}
    \caption{Overview of the training sets featuring the number of cases, mesh sizes and element counts.\label{table:num_elements_dataset}}
\end{table}

\begin{table}[h!]
    \centering
    \renewcommand{\arraystretch}{1.5} 
    \setlength{\tabcolsep}{8pt}      
    \begin{tabular}{crlrr}
        \hline
          & No. & Location & Size & \#Elements \\
        \hline
        \multirow{13}{*}{\rotatebox[origin=c]{90}{Few-shot training}} & 1 & ICA-PCom & 4.3\,mm & 1.0M \\ 
        &  2 & ICA-MCA   &  6.9\,mm & 3.3M \\ 
        &  3 & ICA-AChA  &  8.2\,mm & 3.0M \\ 
        &  4 & ICA-PCom  & 15.9\,mm & 2.2M \\
        &  5 & ICA-MCA   & 10.8\,mm & 1.2M \\
        &  6 & ICA-OA    &  5.7\,mm & 1.1M \\
        &  7 & ICA-OA    &  6.5\,mm & 1.9M \\
        &  8 & ICA-MCA   &  4.5\,mm & 2.0M \\
        &  9 & MCA M1-M2 &  5.5\,mm & 1.8M \\
        & 10 & MCA M1-M2 &  7.9\,mm & 2.7M \\
        & 11 & ICA-PCom  &  5.6\,mm & 3.4M \\
        & 12 & ICA-PCom  &  4.3\,mm & 2.0M \\
        & 13 & ICA-PCom  &  5.7\,mm & 1.6M \\
        \hline
        \multirow{4}{*}{\rotatebox[origin=c]{90}{Validation}} & 14 & MCA M1 & 4.7\,mm & 2.2M \\
        & 15 & MCA M1-M2 & 4.4\,mm & 2.8M \\
        & 16 & MCA M1-M2 & 6.9\,mm & 2.0M \\ 
        & 17 & VA        & 5.0\,mm & 1.8M \\ \hline
    \end{tabular}
    \caption{Summary of cases used in the few-shot training and validation including the aneurysm locations, }
\end{table}

\subsection{Solver methodology}
\label{app-subsec-solver}
\begin{figure}[htb]
    \centering
    \includegraphics[width=\linewidth]{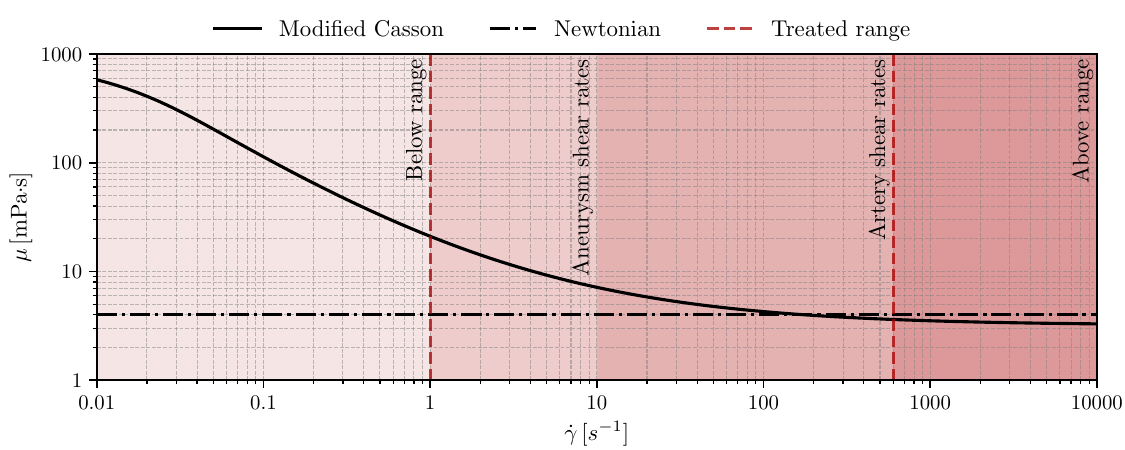}
    \caption{Blood viscosity over the shear rates $\dot{\gamma}$ with the inscribed ranges found in brain aneurysm hemodynamics. The modified Casson model\,\cite{Suzuki2021} is plotted showcasing the differences in dynamic viscosity with the otherwise commonly used Newtonian assumption of $\mu$\,=\,4\,mPa$\cdot$s.}
    \label{fig:visco}
\end{figure}

\paragraph{Blood rheology}
Hemodynamics in vessels of scales in the range of millimetres are governed by the incompressible Navier-Stokes. In this range scale, the red blood cells suspended in plasma can be modelled as a continuum with shear-thinning properties \cite{Nader2019}. This drop in flow resistance is induced through the agglomeration and the alignment of the disk like erythrocytes, which manifest generally for shear rates $\dot{\gamma}$ at and below those found in brain arteries (see \autoref{fig:visco}). The choice between Newtonian and shear thinning models is not clearly answered in the simulation community, with advocates of both strategies debating over it relevance compared to other sources of uncertainty \cite{Berg2019rev, Khan2017, Xiang2012}. In our study, we employ the modified Casson model\,\cite{Suzuki2021} expressed through \autoref{eq:casson}, since some of the slow recirculations found in the aneurysms could be affected by erythrocyte alignment. A hematocrit (Hct) value of 40\% and a model constant $m$\,=\,100 are employed (see \autoref{fig:visco}).
\begin{equation}
    \begin{split}
        \mu &= \left( \sqrt{\frac{\tau_0}{\dot{\gamma}}\left( 1 - \exp{(-m\dot{\gamma})} \right)} +\sqrt{\mu_0} \right)^2\\[0.5em]
        \tau_0 &= (0.888\,\text{Hct} - 23.753) \,\text{mPa}\\[0.5em]
        \mu_0 &= (0.073\,\text{Hct} + 0.599) \,\text{mPa$\cdot$s}
    \end{split}
    \label{eq:casson}
\end{equation}

\paragraph{Navier-Stokes equations}
The solution to the incompressible Navier-Stokes is approximated by an in-house Finite Element solver for linear elements. To handle both convection-dominated flows and the inherent incompatibility of same-order interpolations for velocities $\vec{u}$ and pressure $p$, which is included in the Cauchy stress tensor $\sigma$, a Variational Multiscale-type stabilization\,\cite{Hachem2010} was implemented to solve the weak form shown in \autoref{eqn:navierstokesVMS}. The algebraic subgrid scales $\vec{u}'$ and $p'$ are based on the residuals of the momentum and mass conservation equations\,\cite{Codina2002} as well as some stabilization parameters obtained from Fourier analysis\,\cite{Tezduyar2000}. The time derivative in the momentum equation is discretized with a second-order semi-implicit differentiation formula using 1\,ms timesteps to obtain Courant–Friedrichs–Lewy numbers in the order of 1. To wash out the initial transient at least two cardiac cycles were run for each case, considering only the last cycle for further processing.

\begin{equation}
	\begin{split}
		\int_\Omega \underbrace{\rho\left(\partial_t\vec{u}+(\vec{u}\cdot\nabla)\vec{u} - \vec{g} \right)\cdot \vec{w} + \sigma:\nabla\vec{w}}_{\text{Galerkin terms}} ~d\Omega \, +\\
		+ \int_\Omega \underbrace{\rho(\vec{u}\cdot\nabla)\vec{w}\cdot \vec{u}'}_{\text{Upwind stabilization}} + \underbrace{p'\nabla\cdot\vec{w}}_{\text{Momentum divergence}}~d\Omega
		= \int_{\Gamma} \underbrace{\vec{w}\cdot\sigma\cdot\vec{n}}_{\text{Boundary stress}}~d\Gamma \\
		\int_\Omega \underbrace{q\nabla\cdot\vec{u}}_{\text{Galerkin term}} + \underbrace{\vec{u}'\cdot\nabla q}_{\text{Pressure stabilization}} ~d\Omega = 0
	\end{split}
	\label{eqn:navierstokesVMS}
\end{equation}

\begin{figure}[htb]
    \centering
    \includegraphics[width=0.8\linewidth]{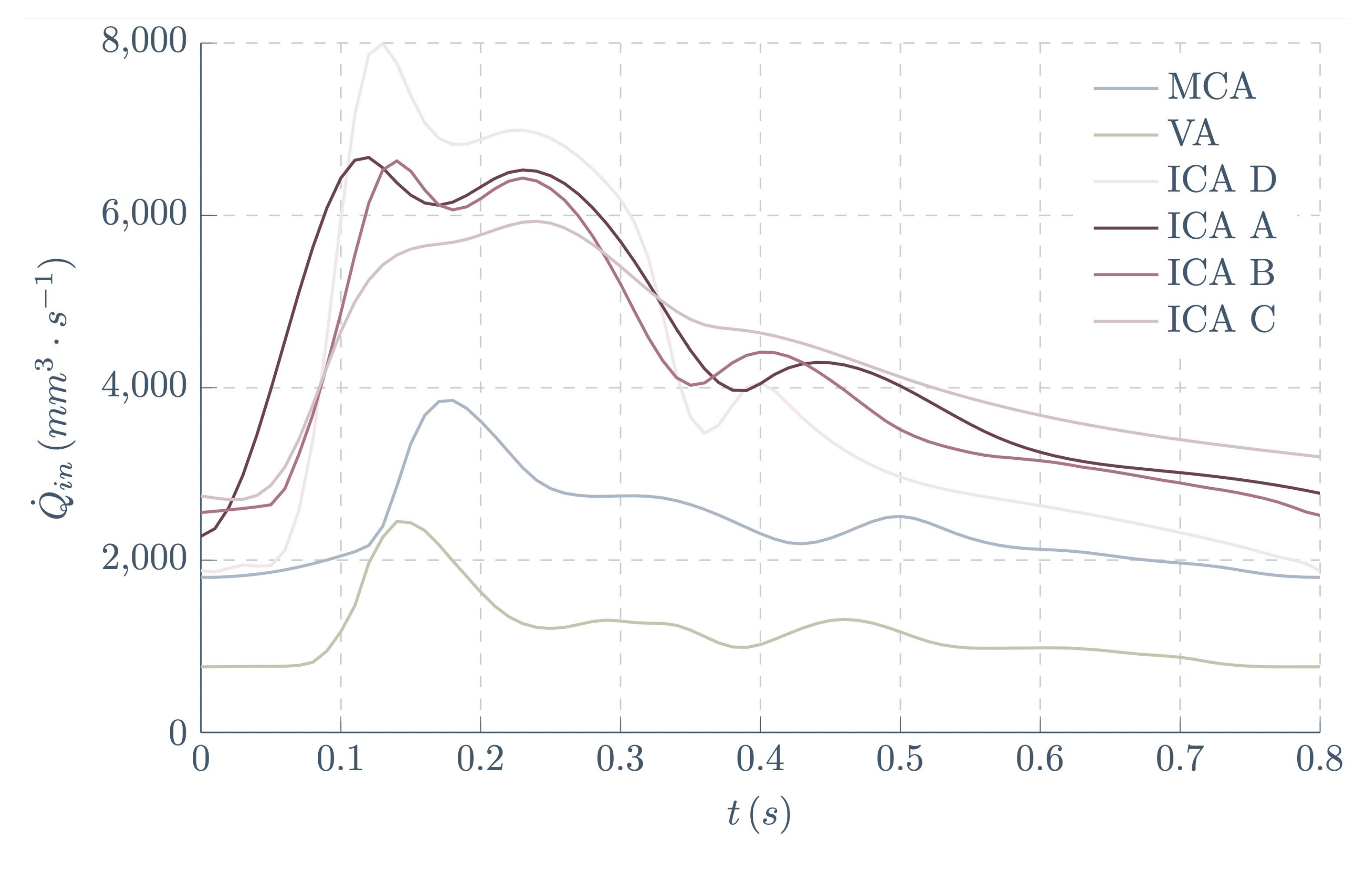}
    \caption{Waveforms used in the fewshot and validation cases. Four variations of the ICA waveform (A-D)\,\cite{Hoi2011, Najafi2021} are included to cover characteristic properties for older and younger patients, as well as different period times. The waveforms for the smaller VA and MCA\,\cite{Reymond2009} are less reliable due to the technical difficulties in acquiring them and are therefore used only once.}
    \label{fig:waveforms}
\end{figure}

\paragraph{Boundary conditions}
The domain boundaries were divided into three categories: inlets, outlets, and arterial walls. The inflow was determined using waveforms obtained via cine phase-contrast magnetic resonance imaging for each parent vessel: the ICA\,\cite{Hoi2011, Najafi2021}, the MCA\,\cite{Reymond2009}, and the VA\,\cite{Reymond2009}. To account for patient variability and leverage the ICA's relatively high flow resolution, four distinct waveforms representing different age groups and health conditions were employed. The flow rates were translated into parabolic inflow profiles with transient magnitudes\,\cite{Valen-Sendstad2018}. An overview of the different waveforms is available in \autoref{fig:waveforms}.
Outflows were treated with traction-free boundary conditions, and the walls were modeled as rigid, no-slip boundaries following the literature\,\cite{Berg2019rev}.

\paragraph{Hemodynamic indicators}
\label{para:metrics-details}
A decisive factor in the initiation and growth of intracranial aneurysms is though to be the Wall Shear Stress (WSS); a measure of the friction exerted by the flow onto the arterial walls\,\cite{Urschel2021, Zimny2021, Meng2014}. Abnormally high WSS and positive WSS gradients have been associated with aneurysm initiation and the development of nearly translucent aneurysm tissue\,\cite{Zimny2021}. Abnormally low and oscillatory WSS on the other hand is observed to be present in aneurysms with atherosclerotic walls\,\cite{Meng2014}. In technical terms, the vectorial WSS $\vec{\tau}_{\text{wss}}$ is the tangential component of the stress tensor $\sigma$ acting at the arterial boundary. It can be expressed with the help of the surface normal $\vec{n}$ as shown in \autoref{eq:wss}.

\begin{equation}
    \vec{\tau}_{\text{wss}} = \vec{n} \times (\sigma\cdot\vec{n})\times \vec{n}
    \label{eq:wss}
\end{equation}

The WSS is commonly averaged in time (TAWSS) to facilitate its analysis\,\cite{Li2019, Xiang2014}. Its expression is given in \autoref{eq:tawss} with $T$ denoting a cardiac cycle.
\begin{equation}
    \text{TAWSS} = \frac{1}{T}\int_T ||\vec{\tau}_{\text{wss}}||\,dt
    \label{eq:tawss}
\end{equation}
The TAWSS encodes overall high and low shearing regions and is more robust to the numerical framework than single WSS frames. The oscillatory components of the shearing previously mentioned are condensed into the Oscillatory Shear Index (OSI). This indicator, given by \autoref{eq:osi}, returns values close to 0 for unidirectional shearing, and will approach 0.5 if the shearing direction is inverted during the cardiac cycle. Despite of its wide-spread use in the analysis of cardiovascular diseases, the OSI suffers more from numerical inaccuracies and other modelling choices\,\cite{Ahn2014}, an the record of maximal values is unclear in regards to the size of the patches over which an elevated quantity should be considered.
\begin{equation}
    \text{OSI} = \frac{1}{2} \left( 1 - \frac{\left\| \int_{T} \vec{\tau}_{\text{wss}} \, dt \right\|}{\int_{T} \|\vec{\tau}_{\text{wss}}\| \, dt} \right)
    \label{eq:osi}
\end{equation}
Apart from the OSI and TAWSS, there are numerous other WSS-derived quantities denoting other properties of the shear stress\,\cite{Chung2018}. These however often lack broad multi-center studies confirming their relationship with IA stability.

A last factor brought into discussion in more recent studies is the characterization of the flow as stable or chaotic. While classifying the flow into either of the categories can be biased, this indicator has received much attention in cases of rupture of treated aneurysms\,\cite{Li2019}, as well as larger longitudinal studies\,\cite{Cebral2017}. In both studies, a significant amount of ruptured or unstable aneurysms featured complex flows including recirculation zones and secondary vortices.

\subsection{Inputs and Outputs}
\label{appendix:inputs}
Each mesh in our datasets contains a vector of size 15 on each node, constituting the features. They were built using the method detailed in \autoref{app-subsec-solver}. 

\begin{itemize}
    \item the velocity $\vec{u} \in \mathbb{R}^3$
    \item the acceleration $\vec{a} \in \mathbb{R}^3$, computed between the current and the previous step
    \item the node position $\vec{x} \in \mathbb{R}^3$
    \item the geometric distance to the inflow plane $d \in \mathbb{R}$
    \item the norm of the velocity $\lVert \vec{u} \rVert \in \mathbb{R}$
    \item the mean, minimum and maximum velocity values at the inflow $(\bar{m}, m, M) \in \mathbb{R}^3$
    \item the node type (if the node is part of the inflow, outflow, artery wall or inside the artery) $n \in \mathbb{N}$
\end{itemize}

The only outputs we try to predict is the velocity $\vec{u} \in \mathbb{R}^3$ at the following time step. The model itself does not predict $\vec{u}_{t+\Delta t}$ directly, but rather $\vec{a}_{t+\Delta t}$ and then sums its prediction with the current velocity.

\subsection{Dataset size}
\label{appendix:dataset-size}

We compute each dataset's size in terms of training tokens by multipling the average number of nodes per mesh, the number of time steps per trajectory, and finally the number of geometry. Following this strategy, we obtain 64M tokens for the coarse version of AnXplore and 1.6B tokens for its fine version. Similarly, we obtain 42M tokens for the coarse version of the fine-tuning dataset and 1.25B tokens for its fine version. 

\subsection{Clinical data management}

\begin{figure}[htb]
    \centering
    \includegraphics[width=0.9\linewidth]{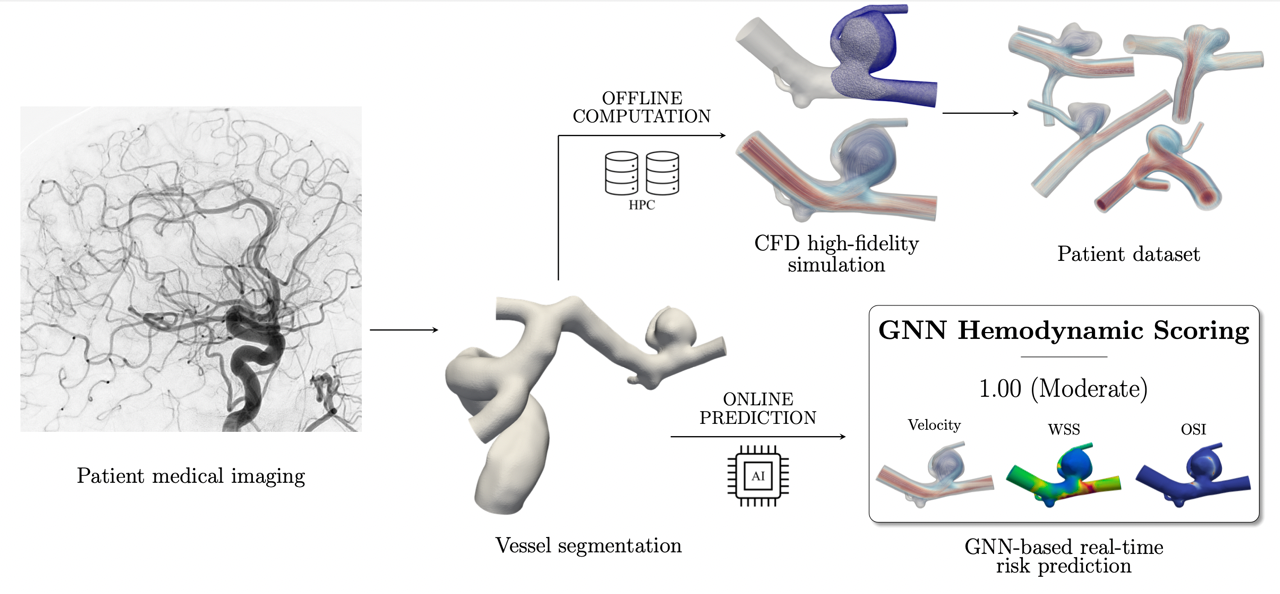}
    \caption{Data management pipeline for GNN clinical inference and patient-specific dataset.}
    \label{fig:data_pipeline}
\end{figure}

All patient-specific cases acquired directly from medical imaging were segmented to generate three-dimensional numerical meshes representing the vascular domain surrounding the aneurysm. From these cases, real-time GNN predictions can be computed for clinical assessment, while extended CFD simulations are performed to continuously expand the patient dataset (\autoref{fig:data_pipeline}). 

\section{Model}\label{app:sec:model}
\subsection{Architecture}

We consider a mesh as an undirected graph $\mathcal{G} = (\mathcal{V},\mathcal{E})$. 
$\mathcal{V} = \{\mathbf{x}_i\}_{i=1:N}$ is the set of nodes, where each $\mathbf{x}_i \in \mathbb{R}^{p}$ represents the attributes of node $i$.
$\mathcal{E} = \{\left(\mathbf{e}_k, r_k, s_k\right)\}_{k=1:N^e}$ is the set of edges, where each $\mathbf{e}_k$ represents the attributes of edge $k$, $r_k$ is the index of the receiver node, and $s_k$ is the index of the sender node. 

In the remaining of our architecture, we omit the attributes of the edges and consider each node as a token. We note $\mathbf{X} = (\mathbf{x}_1, \mathbf{x}_2,...,\mathbf{x}_N)^{T} \in \mathbb{R}^{N \times p}$ our input matrix, made of $N$ tokens of dimension $p$. Let $\mathbf{Z} = (\mathbf{z}_1, \mathbf{z}_2,...,\mathbf{z}_N)^{T} \in \mathbb{R}^{N \times d}$ be the $d$-dimensional embedding of our nodes. We define $\mathbf{A}$ as the adjacency matrix of our graph, setting $\mathbf{A}_{ij} = \mathbf{A}_{ji} = 1$ if $(i,j) \in \mathcal{E}$ and $0$ otherwise. \footnote{It is important to note that: (1) we do not use self loops, so $\mathbf{A}_{ii} = 0, \forall i$ and (2) Even if 2 nodes can be linked by 2 distinct paths, we keep their adjacency coefficient to 1. This is similar to the usual normalized adjacency matrix without self-loop used in Graph Convolutional Networks (GCN) $(\mathbf{D}+\mathbf{I})^{-1/2}\mathbf{A}(\mathbf{D}+\mathbf{I})^{-1/2}$ where $\mathbf{D}$ is the diagonal degree matrix.}

Our model follows an Encode-Process-Decode architecture similar to \cite{Battaglia2018}. The Encoder maps the input nodes into a latent space. We then apply $L$ layers of our transformer architecture. Finally, the Decoder maps back our outputs into a meaningful space.

At each step, our model is auto-regressive, meaning that the output of our model is used as an input for the next step of the simulation.

\subsection{Encoder and Decoder}

We simplify the architecture from \cite{pfaff2021learning} by using only two linear layers to encode our inputs. We also use only a node encoder since our model does not use edge attributes. Our encoder maps our nodes $\mathbf{X} \in \mathbb{R}^{N \times p}$ into a latent space $\mathbf{Z} \in \mathbb{R}^{N \times d}$. The parameter $d$ is shared across all our layers as the main width parameter.

\begin{figure}[!ht]
    \centering
    \includegraphics[width=\linewidth]{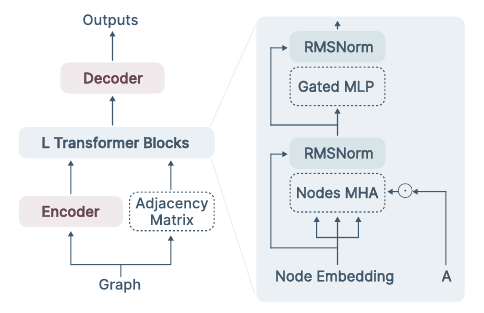}
    \caption{\small\textbf{The Masked Transformer architecture.} \textbf{(left)} Our model takes the graph nodes' features and the adjacency matrix as inputs. The adjacency matrix is improved with Dilation, Global Attention, and Random Jumpers. \textbf{(middle)} Each transformer is made of masked Multi-Head Self-Attention followed by a Gated MLP with residual connection and Layer Normalization. \textbf{(right)} The Multi-Head Self-Attention masks the node's features by computing a sparse matrix multiplication indexed on the Augmented Adjacency matrix. The Gated MLP processes the features in two branches with an expansion factor $e$.}
    \label{fig:model-archi}
\end{figure}

The Decoder generates an output from the latent space using two linear layers. An overview of the architecture is available in \autoref{fig:model-archi}.

\subsection{Processor}

Our Processor is made of $L$ Transformer Blocks. Each block takes the latest latent representation $\mathbf{Z}$ and the Adjacency Matrix $\mathbf{A}$ as input. Each block has two sub-layers: a Masked Multi-Head Self-Attention layer and a Gated MLP layer. We also add residual connections around those two layers, following the original Transformer implementation from \cite{Vaswani2017}. We follow those connections with Layer Normalization using RMSNorm \cite{zhang2019rootmeansquarelayer}. 

We can summarize our architecture as follows:

\begin{align}
    \mathbf{Z}_0 &= \text{MLP}(\mathbf{X}) && \mathbf{X} \in \mathbb{R}^{N \times p}, \mathbf{Z} \in \mathbb{R}^{N \times d} \label{eq:embedding} \\
    \mathbf{Z'}_l &= \text{RMSNorm}\big(\text{MMHA}(\mathbf{Z}_{l-1}, \mathbf{A}) + \mathbf{Z}_{l-1}\big) && \ell \in [1\ldots L] \label{eq:msa_apply} \\
    \mathbf{Z}_l &= \text{RMSNorm}\big(\text{GatedMLP}(\mathbf{Z'}_{l}) + \mathbf{Z'}_{l}\big) && \ell \in [1\ldots L] \label{eq:mlp_apply} \\
    \mathbf{y} &= \text{MLP}(\mathbf{Z}_L) \label{eq:final_rep}
\end{align}

We used a sparse representation of the adjacency mask to implement each transformer layer using \text{DGL}:

\begin{lstlisting}[language=Python]
def scaled_dot_product_attention(
   q: Tensor,
   k: Tensor,
   v: Tensor,
   att_mask: SparseMatrix,
) -> Tensor:
    scaling_factor = k.size(-1) ** 0.5
    q = q / scaling_factor
    attn = dglsp.bsddmm(att_mask, q, k.transpose(1, 0))
    attn = attn.softmax()
    return dglsp.bspmm(attn, v)
\end{lstlisting}

\subsection{Augmented Adjacency Matrix}

The original transformer has a complexity of $O(n^2)$ (with $n$ being the sequence length, or the number of nodes in our case) with every node attending to every node. We move from that paradigm by using a natural formulation when working with graphs and use the Adjacency Matrix as a mask. This is similar to using a Sliding Window of width $w$ with $w$ the maximum degree $\textit{deg}_{max}$ of our graph. This leads to a complexity of $O(n\times \textit{deg}_{max})$.

\begin{figure}
  \centering
  \includegraphics[width=\columnwidth]{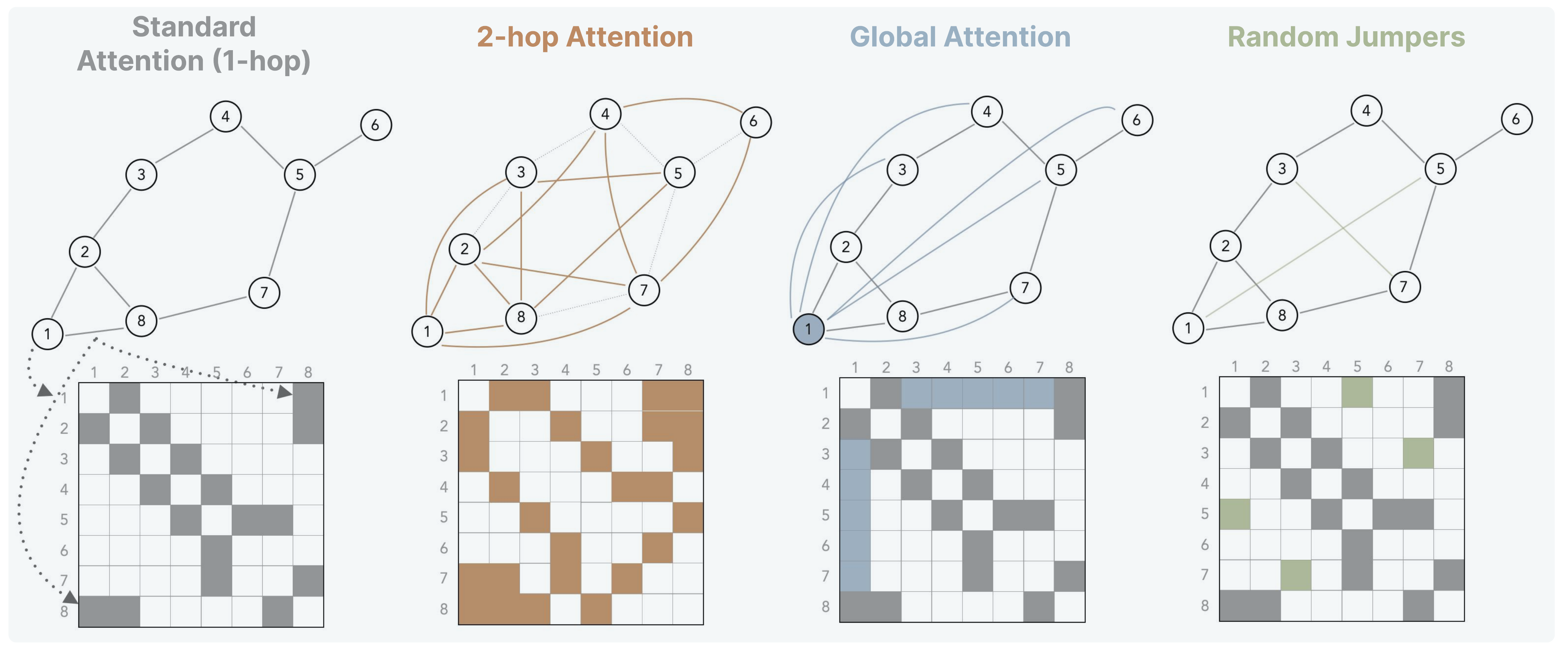}
  \caption{\small\textbf{Details of the Augmented Adjacency matrices} \textbf{(1)} Default Adjacency Matrix $A$. \textbf{(2)} Adjacency Matrix with a Dilation of size $2$. \textbf{(3)} Adjacency Matrix with Global Attention based on node $1$. \textbf{(4)} Adjacency matrix with Random Jumpers between nodes $(1,5)$ and $(3,7)$.}
  \label{fig:adjacency}
\end{figure}

We go one step-further and take inspiration from \cite{garnier2024transformer, zaheer2021bigbirdtransformerslonger} by building an augmented Adjacency Matrix (see \autoref{fig:adjacency}) with random connection, dilatation and global attention.

\paragraph{Dilated Adjacency}

To increase the number of neighbors seen by each node, one can increase the size of the $K$-hop, as seen above. In order to increase the receptive field in a similar fashion without increasing the number of edges too much, one can define a Dilated Adjacency Matrix. Instead of using $\mathbf{A}$, we chose to use $\mathbf{A}^k$. Similar to \cite{beltagy2020longformerlongdocumenttransformer}, we let our model focus on different sizes of receptive fields by setting different adjacency matrices for different heads. We use \emph{2-Dilation}: $\mathbf{A}$ for layers $0$ to $9$, $\mathbf{A}$ for layers $10$ to $14$ on half of the heads and $\mathbf{A}^2$ on the remaining heads.

\paragraph{Adjacency with Random Connections}

To address the inherent locality of updates in our architecture, we introduce random edges to enhance the geometrical length of information flow. For a given number $j$, we randomly select $j$ pairs of nodes and add an edge between them. This approach enables information to propagate much further at a very low cost.

\paragraph{Adjacency with Global Attention}

Similar to how certain words hold more significance than others in a task, some nodes are more important than others within a graph. To address this, we introduce Global Attention by symmetrically connecting specific nodes to every other node in the Adjacency Matrix. We connect 5\% of the Inflow Nodes to every other node in the graph.

\subsection{Pretraining technique}

We always split the training phase in two: a first half where we mask regions of the mesh, and a second half where every nodes are available. To achieve this, we use an AutoEncoder architecture where both the Encoder and the Decoder are trained with masked nodes, and where only the Encoder is fined-tuned with every nodes available. 

The aim is to make the next-step prediction (or the reconstruction) of a mesh given a partially visible input. We follow the strategy of \cite{he2021masked} by having an asymetric architecture, meaning our Encoder is much larger than our Decoder. This does not increase the training budget since most of it is spent during pre-training, where the Encoder inputs are mostly masked.

\begin{figure}[!ht]
    \centering
    \includegraphics[width=0.88\textwidth]{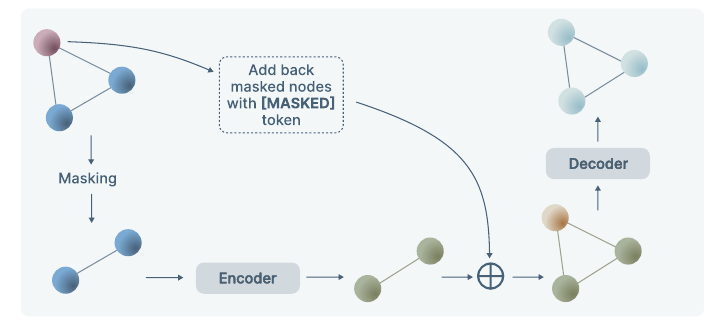}
    \caption{The overall AutoEncoder architecture. More details can be found in \cite{garnier2025meshmask}}
    \label{fig:mae-archi}
\end{figure}

At each training step, we randomly sample a fraction of the existing nodes in a given mesh. We then proceed to completely remove them from the mesh, including all edges that are connected to at least one masked node. We display in \autoref{fig:mae-archi} a brief overview of this architecture. 

After being passed to the Encoder, we reconstruct the initial mesh by:
\begin{itemize}
    \item Replacing hidden nodes by a shared learnable {\tt [MASKED]} token (following \cite{devlin2019bert})
    \item Replacing other nodes by their predicted values from the Encoder
    \item Rebuilding hidden edges with geometric data
\end{itemize}

\subsection{Parameters selection}

\paragraph{Ablation Study}

We follows the ablation study from \cite{garnier2024transformer} and select the following key parameters:
\begin{itemize}
    \item We use $L=15$ transformer blocks
    \item Each block uses a latent dimension $d=512$
    \item We use an expansion factor of size $e=3$ for the Feed Forward Network. This leads to a forward dimension of size $1536$
    \item The Adjacency matrix is augmented with 20\% of random edges, 5\% of global nodes
    \item The matrix is also augmented using a dilation of size 2, of half the heads of the 5 last transformer blocks
\end{itemize}

\begin{figure*}[tbh!]
  \centering
  \includegraphics[width=0.79\textwidth]{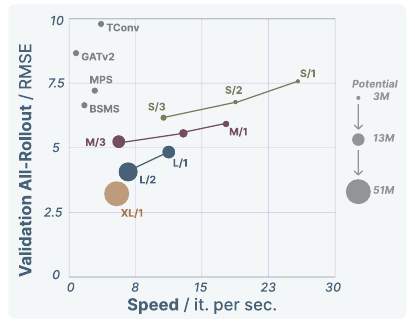}
  \caption{\small\textbf{RMSE over autoregressive trajectories on the Coarse Aneurys Dataset.} Performances are shown without any masking pre-training and with a standard Adjacency Matrix without augmentations. Buble area indicates the Potential of each model (\# parameters $\times$ size of $K$-hop.) Performances increase with the size of the model and the size of the hop. Our new architecture constantly beats the previous SOTA based on message-passing architecture. Our \textbf{S/1} model achieves similar performances while being $7\times$ faster. Our \textbf{XL/1} model achieves an RMSE $3$ times lower.}
  \label{fig:overview}
\end{figure*}

An overview of the results different models sizes is available in \autoref{fig:overview}.

\paragraph{Scaling Law}

We investigate how to balance training steps and number of parameters for a given fixed FLOPs budget. To study this, we model the final All-Rollout RMSE as a function of $P$ the number of parameters and $D$ the number of training nodes.

We conduct extensive training on various models using the coarse AnXplore dataset, focusing on six different FLOPs budgets. Each training run adheres to a cosine decay schedule that corresponds to the required number of training steps, ensuring that all comparisons are consistent across runs.

For each specified FLOPs budget, we plot the parameters of the model that achieves local minima and fit a power law to the data (refer to \autoref{fig:scalinglaws}). Our analysis reveals that \(N \propto C^{0.75}\), where \(C\) represents the FLOPs budget. This exponent is notably higher than the findings in \cite{hoffmann2022trainingcomputeoptimallargelanguage}, yet aligns closely with the results reported by \cite{kaplan2020scalinglawsneurallanguage}.

In \autoref{fig:scalinglaws}, we emphasize the sizes of the models analyzed in this study along with those discussed in the aforementioned papers. Both \cite{kaplan2020scalinglawsneurallanguage} and \cite{hoffmann2022trainingcomputeoptimallargelanguage} focus on pretraining Large Language Models (LLMs), with model sizes ranging from 300k to 700M parameters and from 100M to 10B parameters, respectively. Our models fall within the range of 50k to 50M parameters. These studies also resulted in power law fits, showing \(N \propto C^{0.73}\) and \(N \propto C^{0.50}\), respectively.

\begin{figure}[!h]
  \centering
  \includegraphics[width=0.79\textwidth]{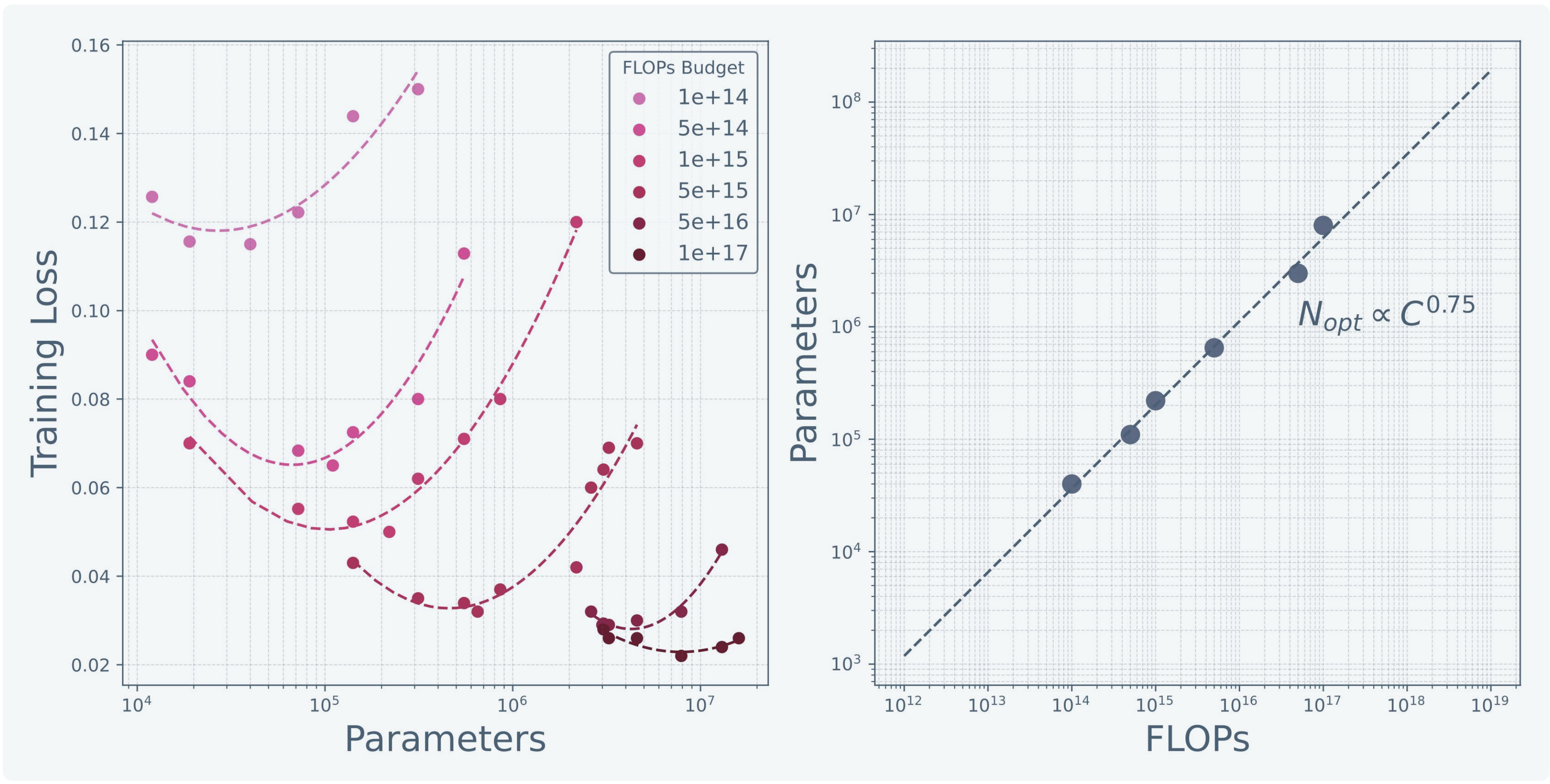}
  \caption{We train various models for a specific FLOPs budget, adjusting the number of iterations to match the cosine cycle length. Each isoFLOPs curve is analyzed to identify a local minimum. \textbf{(middle)} We then plot the model corresponding to the local minimum of each curve and present the power law estimation.}
  \label{fig:scalinglaws}
\end{figure}

\begin{figure}[!h]
  \centering
  \includegraphics[width=0.79\textwidth]{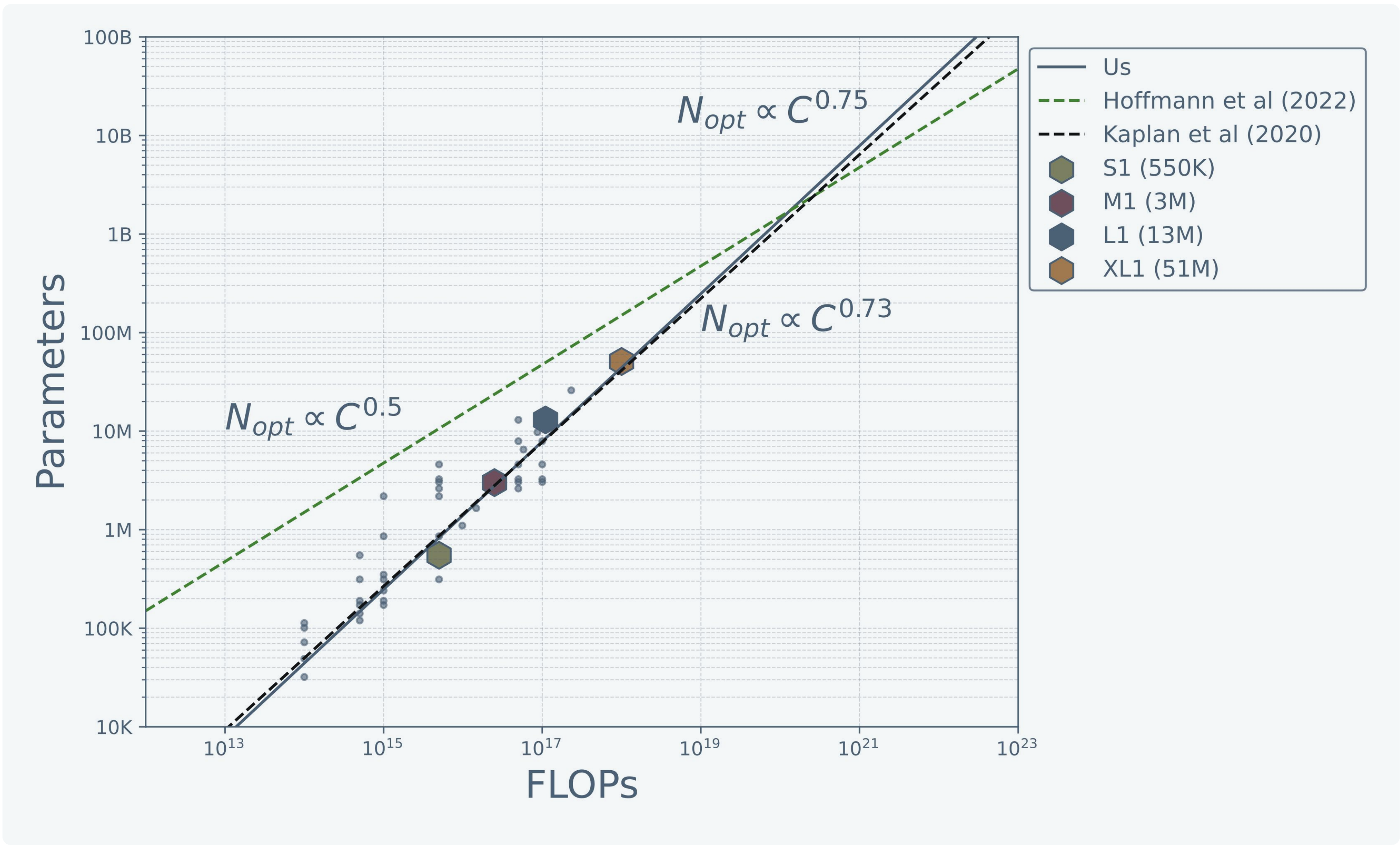}
  \caption{Our predictions are compared with those from Hoffman and Kaplan. The models we trained are marked in blue, while the gray area represents the range of models studied by \cite{kaplan2020scalinglawsneurallanguage}, and the green area shows the results from \cite{hoffmann2022trainingcomputeoptimallargelanguage}.}
  \label{fig:scalinglaws}
\end{figure}

\section{Training}\label{app:sec:training}
\subsection{Noisy inputs}

Since our model will make autoregressive predictions over long rollouts, it is required to mitigate error accumulations.
Since during both pre-training and finetuning, the model is only presented with steps separated by at most one step $\Delta t$, it never sees accumulated noise
from previous predictions.
To simulate this, we use the same approach as \cite{sanchezgonzalez2020learning} and \cite{pfaff2021learning} and make our inputs noisy.
We add random noise $\mathcal{N}(0,\sigma)$ to the dynamical variables: the velocity and the acceleration. We use the value $\sigma = (10,10, 1)$ for both velocity and acceleration on the AnXplore dataset, and $\sigma = (10,10, 10)$ on all the other training cases. Those values are based on the error distribution of our model without any noise. In the case of aneurysm, since the velocity inflow is different depending on the time steps, the error distribution varies a lot across a trajectory. One solution that was tried but proved to be unsuccessful was to use dynamic noise depending on the time steps, using the error distribution seen above. This led to similar results to the static noise we end up using.

\subsection{Optimisation schedule}

The training was split in 4 different phases:

\begin{enumerate}
    \item We performed 150000 gradient descents with a cosine decay from $10^{-4}$ to $10^{-7}$ on a coarse version of the AnXplore dataset with the masking pretraining technique detailed above
    \item We performed 150000 gradient descents with a cosine decay from $10^{-4}$ to $10^{-7}$ on a coarse version of the AnXplore dataset, finetuning only the Decoder. The Decoder now becomes the main model
    \item We performed 45000 gradient descents with a cosine decay from $10^{-4}$ to $10^{-7}$ on the AnxPlore dataset
    \item We performed 20000 gradient descents with a cosine decay from $10^{-5}$ to $10^{-8}$ on a coarse version of the few-shot dataset
    \item We performed 20000 gradient descents with a cosine decay from $10^{-5}$ to $10^{-8}$ on the few shot dataset
\end{enumerate}

We also compare in differences between an Adam \cite{kingma2017adam} and an AdamW \cite{loshchilov2019decoupledweightdecayregularization} optimizer. We find the AdamW optimizer to perform better and use it in all our training, with parameters $\beta _1 = 0.9, \beta _2 = 0.95$, and $\text{weight\_decay} = 10^{-4}$.

\begin{figure}[!t]
  \centering
  \includegraphics[width=0.99\textwidth]{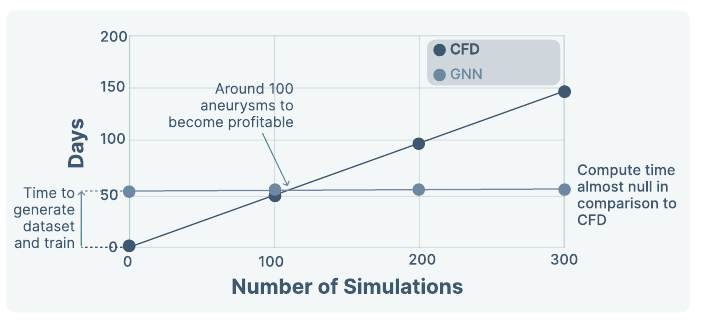}
\caption{
Time spent in days for the training of a \textbf{XL/1} transformer on our datasets, followed by inference simulations. For the groundtruth, we consider time from the \cite{digonnet2007cimlib} solver. For the GNN, we consider the inference time, the training time, as well as the duration from the groundtruth solver to build the datasets. We can see that even when taking everything into account, only 100s of simulations are necessary for the machine learning approach to be worth it.
} 
  \label{fig:datasetsdetails-time}
\end{figure}

\subsection{Training time}

Following the schedule above, the total training of our model took 10 days on a single A100 GPU. 

\begin{figure}[!t]
  \centering
  \includegraphics[width=0.98\textwidth]{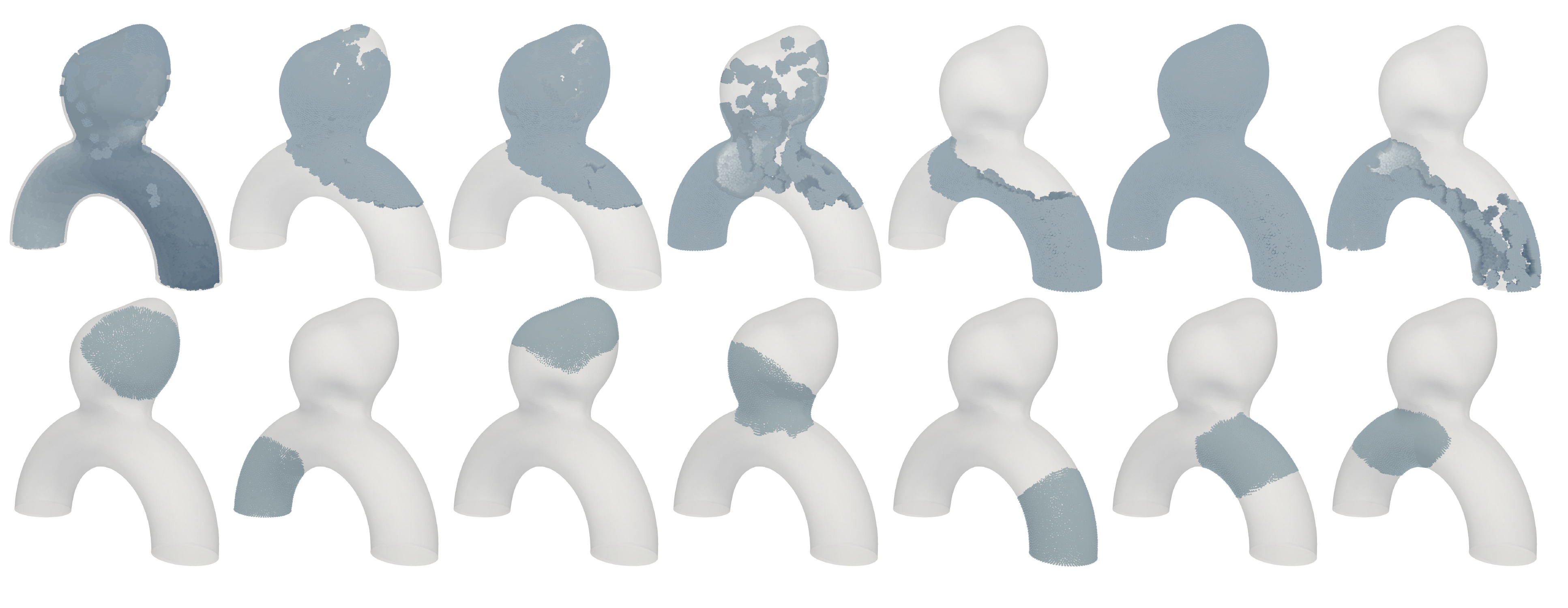}
\caption{
Different sub-mesh generated at each batch update, \textbf{(top)} using random neighbor sampling and \textbf{(bottom)} using METIS.
} 
  \label{fig:submesh}
\end{figure}

Meshes from the real aneurysms shapes were too large to fit in GPU memory and thus necessitated to be split into sub-meshes. The strategy applied was to split one mesh into smaller sub-meshes, and instead of one gradient descent apply multiples ones on each of the sub-mesh. Some sub-meshes can be seen in \autoref{fig:submesh}.

Sub-meshes were generated using two strategies: a random neighbor sampling strategy from \cite{hamilton2018inductiverepresentationlearninglarge} using between 50000 and 100000 random edges. The second strategy used the METIS \cite{Padua2011} algorithm to generate between 7 and 15 disjoint sub-meshes. 

We conducted an extensive study to compare results on model trained with and without this sub-meshs strategy and found no meaningful difference in accuracy. We thus used METIS for most of the training on meshes that were too large.

\section{Evaluation}\label{app:sec:evaluation}

We compute several metrics to evaluate and compare our model.

\subsection{1-step and All-rollout}
\label{appendix:ml-metric}

Our model is trained using a Mean Squared Error loss, average on every node. To evaluate our models, we use the 1-step RMSE and the All-Rollout RMSE defined below:

\begin{align*}
    \text{1-step}(f) := \frac{1}{TN} \sum _{t=1}^{T} \Bigg(\sum _{i \in V} \big(G_t - f(G_{t-1})\big)_i^2 \Bigg)
\end{align*}

\begin{align*}
    \text{All-Rollout}(f) := \frac{1}{TN} \sum _{t=1}^{T} \Bigg(\sum _{i \in V} \big(G_t - \underbrace{f \circ ... \circ f}_\text{t times}(G_0)\big)_i^2 \Bigg)
\end{align*}

where $f$ is the model, $T$ the number of time steps, $N$ the number of nodes and $G_t$ the ground truth graph at time step $t$.

\begin{figure}[!ht]
	\centering
	\includegraphics[width=1\textwidth]{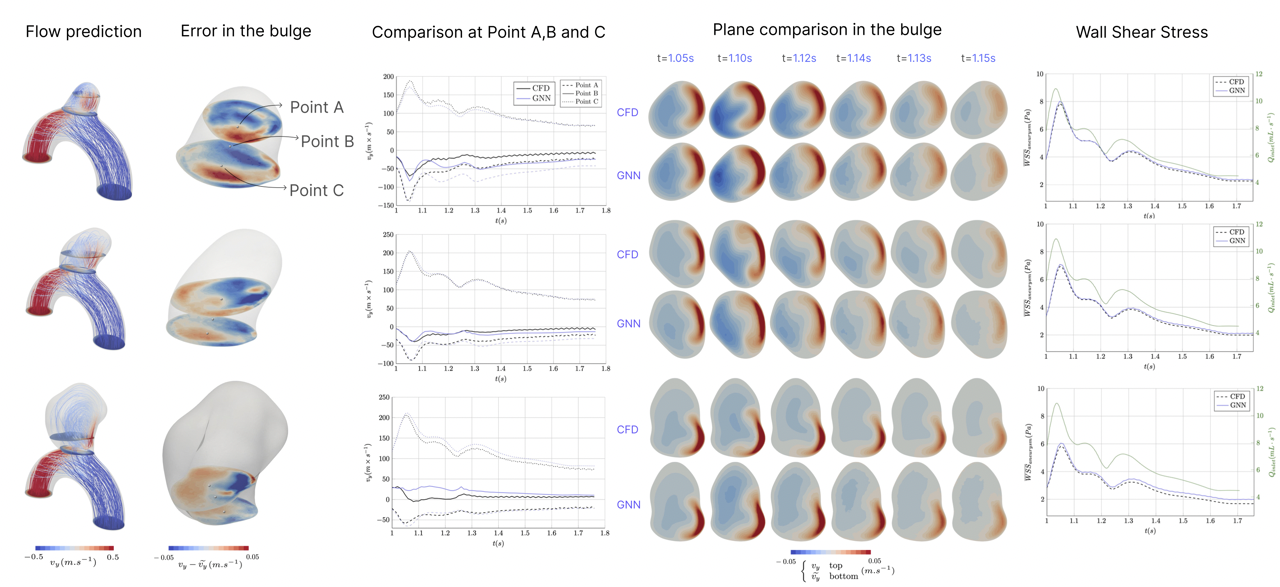}
	\caption{\small
		We showcase results on the pre-training dataset. We display predicted systolic flow, errors inside the bulge at $y=8$ and $y=10$, a comparison of three selected points at $y=8, y=9, y=10$, and a comparison between the CFD (top row) and our transformer (bottom row) for $v_y$ inside the bulge in a 2D plan defined by $x=0, y=10, z=0$. The GNN model accurately captures both velocity fields and derived wall shear features.
	}
	\label{fig:res-pretraining}
\end{figure}

\begin{figure}[!ht]
	\centering
	\includegraphics[width=1\textwidth]{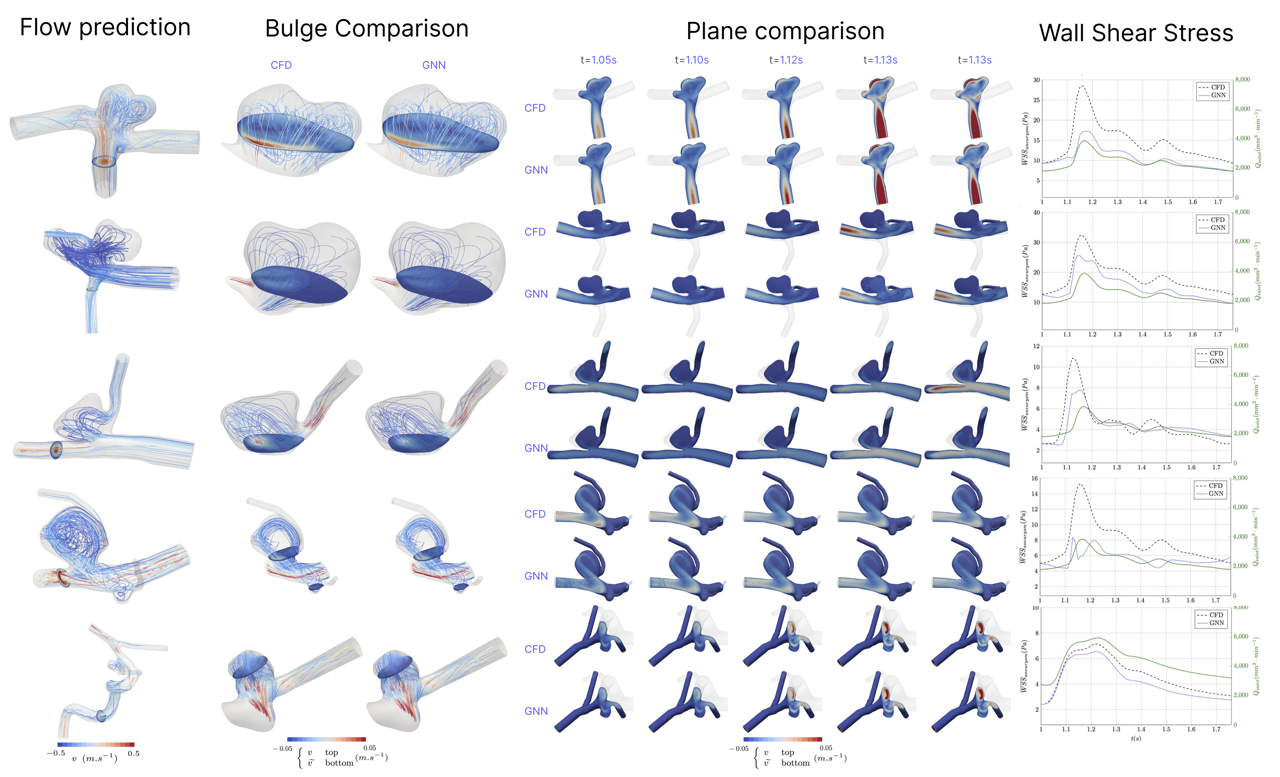}
	\caption{\small
		We showcase results on the validation set and one case from the test set extracted from the fine-tuning dataset. We display predicted systolic flow in the artery and the bulge and a plane comparison inside the shapes between the CFD (top row) and our transformer (bottom row). Overall, flow directions are very well predicted, but our model often lacks velocity magnitude (around 10\% on average). The 4th validation case is also challenging to infer since it has two aneurysms.
	}
	\label{fig:res-validation}
\end{figure}

\subsection{Risk Assessment}
\label{appendix:risk}

Following the work from \cite{Zhou2017, Chung2018, Sheikh2020, Axier2022, Wei2022, Xin2022_CombinationMorphHemodyn, Tian2022}, we synthesize a rule based system using four different metrics: the peak WSS, the TAWSS, the OSI and the peak systolic velocity\footnote{Normalized by the mean inlet velocity.}. We turn each metric into a $\{0,1,2\}$ risk score, directly usable in a logical decision tree. Thresholds are anchored in the cited literature and chosen to be conservative and practical for most CFD pipelines. Our final score is simply the averaged risk score. 

\begin{enumerate}
  \item \textbf{TAWSS (Pa)}
  \[
    \mathrm{risk}_{\mathrm{TAWSS}} =
      \begin{cases}
        1, & \mathrm{TAWSS} \le 1.5~\mathrm{Pa}\quad\text{(pathologically low)}\\
        1, & \mathrm{TAWSS} \ge 6.7~\mathrm{Pa}\quad\text{(high range)}\\
        0, & \text{otherwise.}
      \end{cases}
  \]
  The low–shear cut–off ($\le\!1.5\,\mathrm{Pa}$) is a commonly used pathological threshold linked to endothelial dysfunction and rupture risk \cite{Zhou2017}. The high–shear tail reflects the observation that ruptured lesions often exhibit concentrated jet with elevated shear \cite{Wei2022}. 

  \item \textbf{Peak WSS (Pa)}:
  \[
    \mathrm{risk}_{\mathrm{peakWSS}} =
      \begin{cases}
        2, & \mathrm{peakWSS} \ge 6~\mathrm{Pa} \\
        1, & \mathrm{peakWSS} \ge 4~\mathrm{Pa} \\
        0, & \text{otherwise.}
      \end{cases}
  \]
  In \cite{Xin2022_CombinationMorphHemodyn} , $\mathrm{peakWSS}$ discriminated rupture status with AUC \(=0.681\), but our meta analysis showed that peak WSS is not a very consistent metric for rupture prediction.

  \item \textbf{OSI}
  \[
    \mathrm{risk}_{\mathrm{OSI}} =
      \begin{cases}
        2, & \mathrm{OSI} \ge 0.3 \\
        1, & \mathrm{OSI} \ge 0.15 \\
        0, & \text{otherwise.}
      \end{cases}
  \]
  \cite{Axier2022, Chung2018} reported higher OSI in ruptured aneurysms, with a mean of 0.1 for unruptured ones. The interval from 0.15 to 0.3 is near the upper tail of ruptured values.

  \item \textbf{Peak–systolic bulge velocity}
  \[
    \mathrm{risk}_{V} =
    \begin{cases}
      2, & \mathrm{VR} \ge 80 \\
      1, & \mathrm{VR} \ge 50 \\
      1, & \mathrm{VR} \le 20 \\
      0, & \text{otherwise.}
    \end{cases}
  \]
  While low velocities and recirculated are tied to low WSS, thrombus and well degeneration \cite{Zhou2017, Sheikh2020}, \cite{Chung2018, Sheikh2020} also demonstrates that high and too high velocities are often found in ruptured aneurysms. 
\end{enumerate}






\end{appendices}


\end{document}